\documentclass[lettersize,journal]{IEEEtran}
\usepackage{amsmath,amsfonts}
\usepackage{algorithmic}
\usepackage{algorithm}
\usepackage{booktabs}
\usepackage{array}
\usepackage[caption=false,font=normalsize,labelfont=sf,textfont=sf]{subfig}
\usepackage{textcomp}
\usepackage{stfloats}
\usepackage{url}
\usepackage{verbatim}
\usepackage{graphicx}
\usepackage{cite}
\usepackage{multirow}
\usepackage{xcolor}
\usepackage{makecell}
\usepackage[table]{xcolor}

\begin{document}

\title{Embodied AI: From LLMs to World Models}
\author{Tongtong Feng, Xin Wang,~\IEEEmembership{Member,~IEEE,} Yu-Gang Jiang,~\IEEEmembership{Fellow,~IEEE,} Wenwu Zhu$^{\ast}$,~\IEEEmembership{Fellow,~IEEE}
\thanks{The invited paper}
\thanks{*Corresponding author}
\thanks{Tongtong Feng, Xin Wang and Wenwu Zhu are with the Department of Computer Science and Technology, Tsinghua University, Beijing 100084, China (e-mail: {fengtongtong, xin{\_}wang, wwzhu}@tsinghua.edu.cn). Xin Wang and Wenwu Zhu are also with Beijing National Research Center for Information Science and Technology. Yu-Gang Jiang is with the Institute of Trustworthy Embodied AI, Fudan University, Shanghai 200433, China. (e-mail: ygj@fudan.edu.cn).}
}

% The paper headers
\markboth{IEEE Circuits and Systems Magazine}%
{Shell \MakeLowercase{\textit{et al.}}: A Sample Article Using IEEEtran.cls for IEEE Journals}

\maketitle
\begin{abstract}
Embodied Artificial Intelligence (AI) is an intelligent system paradigm for achieving Artificial General Intelligence (AGI), serving as the cornerstone for various applications and driving the evolution from cyberspace to physical systems. Recent breakthroughs in Large Language Models (LLMs) and World Models (WMs) have drawn significant attention for embodied AI. On the one hand, LLMs empower embodied AI via semantic reasoning and task decomposition, bringing high-level natural language instructions and low-level natural language actions into embodied cognition. On the other hand, WMs empower embodied AI by building internal representations and future predictions of the external world, facilitating physical law-compliant embodied interactions. As such, this paper comprehensively explores the literature in embodied AI from basics to advances, covering both LLM driven and WM driven works. In particular, we first present the history, key technologies, key components, and hardware systems of embodied AI, as well as discuss its development via looking from unimodal to multimodal angle. We then scrutinize the two burgeoning fields of embodied AI, i.e., embodied AI with LLMs/multimodal LLMs (MLLMs) and embodied AI with WMs, meticulously delineating their indispensable roles in end-to-end embodied cognition and physical laws-driven embodied interactions. Building upon the above advances, we further share our insights on the necessity of the joint MLLM-WM driven embodied AI architecture, shedding light on its profound significance in enabling complex tasks within physical worlds. In addition, we examine representative applications of embodied AI, demonstrating its wide applicability in real-world scenarios. Last but not least, we point out future research directions of embodied AI that deserve further investigation.
\end{abstract}

\section{Introduction}

\begin{figure}[t]
\centering
\includegraphics[width=0.45\textwidth]{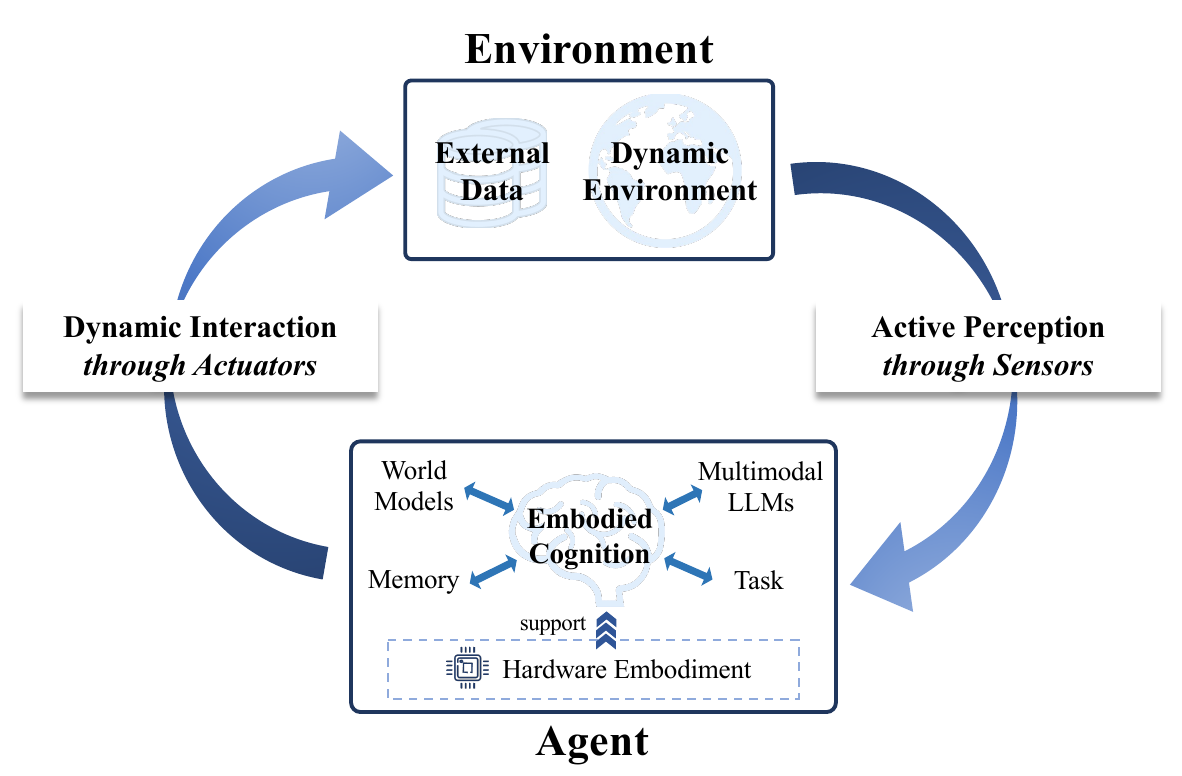}
\caption{The concept of embodied AI.}
\label{fig:concept}
\vspace{-1.3em}
\end{figure}

Embodied Artificial Intelligence (AI) originated from the Embodied Turing Test by Alan Turing in 1950 \cite{turing1950mind}, which is designed to explore whether agents can imitate human intelligence to achieve Artificial General Intelligence (AGI). Among them, agents that only solve abstract problems in digital world (cyberspace) are generally defined as disembodied AI, while those that also can interact with the physical world are regarded as embodied AI. Embodied AI builds on foundational insights from cognitive science and neuroscience \cite{pfeifer2006body, clark1998being}, which claims that intelligence emerges from the dynamic coupling of perception, cognition, and interaction. As shown in Fig. \ref{fig:concept}, embodied AI includes three key components in a closed-loop manner, i.e., 1) active perception (sensor-driven environmental observation), 2) embodied cognition (historical experience-driven cognition updating), and 3) dynamic interaction (actuator-mediated action control). Besides, hardware embodiment \cite{han2015deep, chen2018tvm, jouppi2017datacenter} is also critical due to escalating computational and energy demands, particularly under latency and power constraints of devices in real-world deployment scenarios.

The development of embodied AI has evolved from unimodal to multimodal paradigm. In early stage, embodied AI is primarily studied through focusing on individual components with single modality such as vision, language, or action, where the perception, cognition, or interaction component is driven by one sensory input \cite{engel2014lsd, mur2015orb}, e.g., perception tends to be dominated by the visual modality \cite{macario2022comprehensive}, cognition tends to be dominated by the language modality \cite{liu2023embodied, lu2023llm}, and interaction tends to be dominated by the action modality \cite{jayaraman2018learning, lin2023grounded}. Although these methods perform well within individual components, they are limited by the narrow scope of information provided by each modality and the inherent gaps between modalities across components. The continued development of embodied AI witnesses the limitations of unimodal approaches, promoting a significant shift toward integration of multiple sensory modalities \cite{driess2023palm, jiang2024fisher, wang2024actiview}. As such, multimodal embodied AI \cite{qin2024mp5, yang2025embodiedbench} naturally arises to create more adaptive, flexible, and robust agents capable of performing complex tasks in dynamic environments.

Large Language Models (LLMs) empower embodied AI via semantic reasoning \cite{brohan2022rt} and task decomposition \cite{alayrac2022flamingo, floridi2020gpt}, bringing high-level natural language instructions and low-level natural language actions into embodied cognition. Representative LLM driven works include SayCan \cite{ahn2022saycan}, which i) provides a real-world pretrained natural language action library to constrain LLMs from proposing infeasible and contextually inappropriate actions; ii) uses LLMs to convert natural language instructions into natural language action sequences; and iii) utilizes value functions to verify the feasibility of natural language action sequences in a particular physical environment. These works suggest that LLMs are extremely useful to robots which aim at acting upon high-level, temporally extended instructions expressed in natural language. However, LLMs are only a part of the entire embodied AI system (e.g., embodied cognition), which is limited by a fixed natural language action library and a specific physical environment, making it difficult for LLM driven  embodied AI to achieve adaptive expansion for new robots and environments.

Recent breakthroughs in Multimodal LLMs (MLLMs) \cite{singh2023progprompt, brohan2023rt} and World Models (WMs) \cite{ding2024understanding, ha2018world, lecun2022path} have opened up a new frontier in embodied AI research. MLLMs can act on the entire embodied AI system, bridging high-level multimodal inputting and low-level motor action sequences into end-to-end embodied applications. Semantic reasoning \cite{li2024manipllm, duan2022survey, brophy2023review} leverages MLLMs’ cross-modal comprehension to interpret semantics from visual, auditory, or tactile inputs, e.g., identifying objects, inferring spatial relationships, predicting environmental dynamics. Concurrently, task decomposition \cite{black2410pi0, huang2024rekep, zhao2025cot} employs MLLMs’ sequential logic to break complex objectives into sub-tasks while dynamically adapting plans based on sensor feedback. However, MLLMs often fail to ground predictions in physics-compliant dynamics \cite{liu2024physics} and exhibit poor real-time adaptation \cite{driess2023vla} to environmental feedback.

\begin{figure*}[t]
\centering
\includegraphics[width=\textwidth]{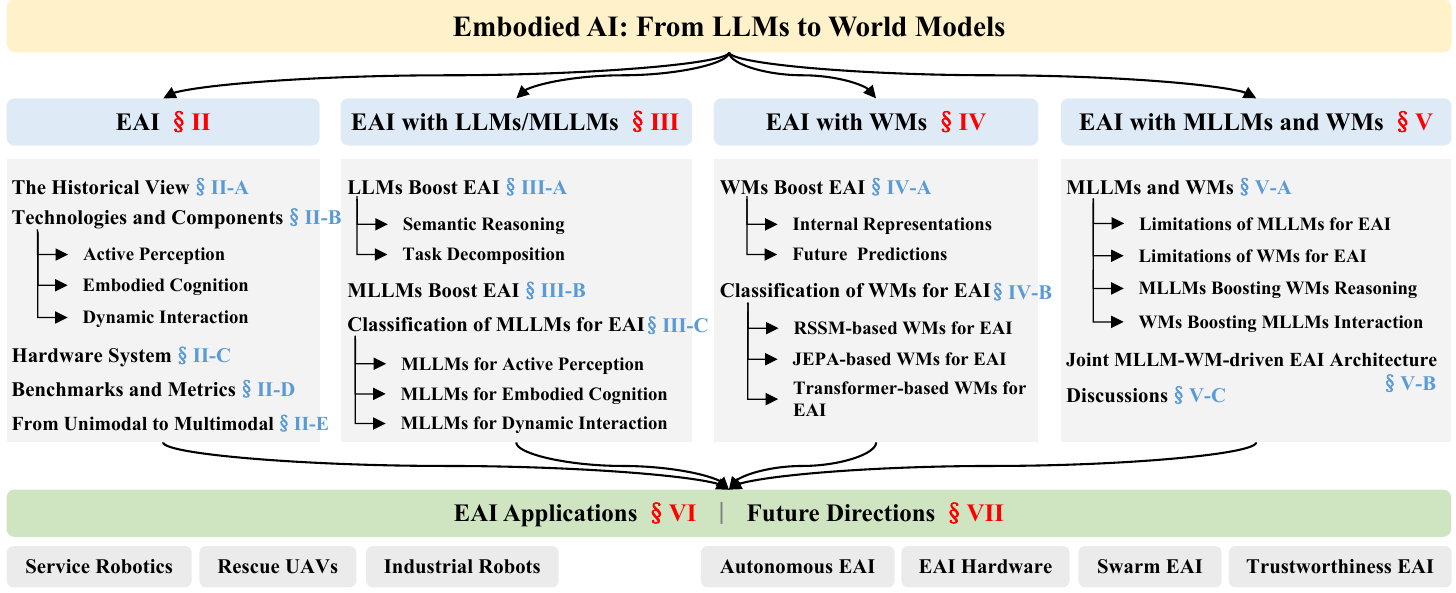}
\caption{This paper comprehensively introduces the basics of Embodied AI (EAI) and the latest advancements of EAI with LLMs/MLLMs and WMs. MLLMs enable contextual task reasoning but overlook physical constraints, while WMs excel at physics-aware simulation but lack high-level semantics. Building upon the above advances, this paper proposes a joint MLLM-WM-driven EAI architecture. Finally, this paper discusss applications and future directions of EAI.}
\label{fig:overview}
\vspace{-1.1em}
\end{figure*}

On the other hand, WMs empower embodied AI by building internal representations \cite{bruce2024genie, chen2022transdreamer, robine2023transformer, wang2024worlddreamer, wu2023paragraph} and making future predictions \cite{haftner2020dream, haftner2021mastering, okada2022dreamingv2, wu2023daydreamer} of the external world. Such WM driven embodied AI is able to facilitate physical law-compliant embodied interactions in dynamic environments. Internal representations compress rich sensory inputs into structured latent spaces, capturing object dynamics, physics laws, and spatial structures, as well as allowing agents to reason about ``what exists'' and ``how things behave'' in their surroundings. Simultaneously, future predictions simulate potential rewards of sequence actions across multiple time horizons aligned with physical laws, thereby preempting risky or inefficient behaviors. However, WM driven approaches struggle with open-ended semantic reasoning \cite{clark2013predictive} and lack the ability of generalizable task decomposition \cite{ha2018world} without explicit priors. 

Building upon the above advances, we further share our insights on the necessity of developing a joint MLLM-WM driven embodied AI architecture, shedding light on its profound significance in enabling complex tasks within physical worlds. MLLMs enable contextual task reasoning but overlook physical constraints, while WMs excel at physics-aware simulation but lack high-level semantics. The joint of MLLM and WM can bridge semantic intelligence with grounded physical interaction. For instance, EvoAgent \cite{feng2025evoagent} designs an autonomous-evolving agent with a joint MLLM-WM driven embodied AI architecture, which can autonomously complete various long-horizon tasks across environments through self-planning, self-reflection, and self-control, without human intervention. We believe that designing joint MLLM-WM driven embodied AI architectures will dominate next-generation embodied systems, bridging the gap between specialized AI agents and general physical intelligence.

We summarize the representative applications of embodied AI as service robotics, rescue UAVs, industrial Robots, and others etc., demonstrating its wide applicability in real-world scenarios. We also point out potential future directions of embodied AI, including but not limited to autonomous embodied AI, embodied AI hardware, and swarm embodied AI etc.

As shown in Fig. \ref{fig:overview}, the rest of this paper is organized as follows. Section \uppercase\expandafter{\romannumeral2} introduces the history, key technologies, key components, and hardware system of embodied AI, discussing the development of embodied AI from unimodal to multimodal angle. Section \uppercase\expandafter{\romannumeral3} presents embodied AI with LLMs/MLLMs, and Section \uppercase\expandafter{\romannumeral4} presents embodied AI with WMs. Section \uppercase\expandafter{\romannumeral5} introduces our insights on designing a joint MLLM-WM driven embodied AI architecture. Section \uppercase\expandafter{\romannumeral6} briefly examines applications of embodied AI. Potential future directions are discussed in Section \uppercase\expandafter{\romannumeral7}.
\section{Embodied AI} 
This section provides a comprehensive overview of embodied AI. We first take a historical view to introduce the development of embodied AI in Subsection \ref{subsec_history}. Based on technological advancements in five foundational areas related to embodied AI, Subsection \ref{subsec_key_tech_comp} and Subsection \ref{hardware} further review the developmental trajectories of core modules in software algorithms and hardware design, respectively. Finally, Subsection \ref{development} discusses an overall analysis of the developmental trends from unimodal to multimodal.

\begin{figure*}[t]
    \centering
    \includegraphics[width=0.9\linewidth]{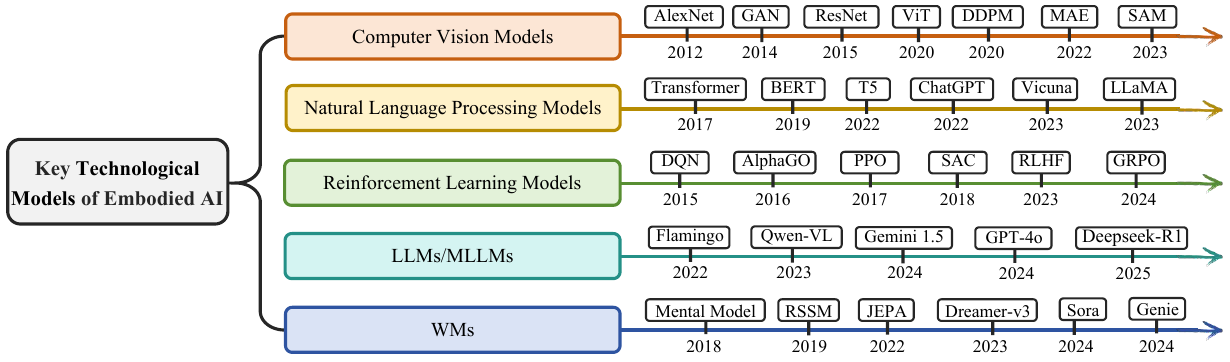}
    \caption{Key technological models of embodied AI. Advancements in Computer Vision (CV) models, Natural Language Processing (NLP) models, Reinforcement Learning (RL) models, LLMs/MLLMs, and WMs have driven progress in embodied AI.}
    \label{fig:2_tech}
    \vspace{-1em}
\end{figure*}

\subsection{The Historical View} \label{subsec_history}
The historical evolution of embodied AI reflects successive transitions from early philosophical foundations to technological breakthroughs in robotics and the rise of learning-driven paradigms, while recent progress in LLMs and WMs is driving an ongoing shift toward the next phase of development. 

The theoretical roots of embodied AI trace to 1950, when Turing introduced the foundational idea that intelligence is inherently linked to physical experience \cite{turing1950computing}. In the 1980s, cognitive science further formalized this view. Lakoff and Johnson emphasized that human cognition arises from bodily experience rather than disembodied symbolic computation \cite{lakoff1980metaphors}, while Harnad’s symbol grounding problem highlighted the necessity of connecting symbolic representations to sensory-motor reality \cite{harnad1990symbol}. Technological advances in robotics during the late 1980s and 1990s brought these ideas into practice. Brooks proposed the subsumption architecture \cite{brooks1986robust, brooks1991intelligence}, promoting behavior-based control through layered, reactive modules grounded in sensorimotor loops. The Cog project \cite{brooks1998cog}  advanced this line by constructing humanoid robots capable of developmental learning, imitation, and social interaction. 
Recently, the success of the learning-driven paradigm has driven the shift in embodied AI from motion control of robots to adaptive interaction \cite{chrisley2003embodied}. In particular, the development of deep learning enables robots to learn complex nonlinear mappings from raw sensor data to action policy, significantly improving navigation and manipulation tasks \cite{levine2016end, mnih2015human}.

While embodied AI has made notable advances, achieving self-reflection intelligence in dynamic, uncertain environments remains a key challenge. Recent progress in LLMs/MLLMs \cite{singh2023progprompt, brohan2023rt} and WMs \cite{ding2024understanding, ha2018world, lecun2022path} have progressively shown promise in overcoming these challenges.

\subsection{The Key Technologies and Components}   \label{subsec_key_tech_comp}
Before discussing the ongoing changes, we systematically review the development of key technologies and components.

\subsubsection{Key Technologies of Embodied AI}
The rapid development of embodied AI is closely tied to advances in foundational technological models such as Computer Vision (CV) models, Natural Language Processing (NLP) models, Reinforcement Learning (RL) models, LLMs/MLLMs, and WMs (as shown in Fig. \ref{fig:2_tech}), which can significantly enhance the capabilities of agents in perception, cognition and interaction.

Specifically, Classic models in computer vision, such as AlexNet \cite{krizhevsky2012imagenet}, GAN \cite{goodfellow2014generative}, ResNet \cite{he2016deep}, ViT \cite{dosovitskiy2020image}, DDPM \cite{ho2020denoising}, MAE \cite{he2022masked}, and SAM \cite{kirillov2023segment} provide the perceptual foundation for embodied agents to interpret high-dimensional sensory inputs in complex environments. In the field of NLP, the evolution from foundational architectures like Transformer \cite{vaswani2017attention}, BERT \cite{devlin2019bert}, and T5 \cite{raffel2020exploring} to large-scale systems such as ChatGPT \cite{achiam2023gpt}, Vicuna \cite{chiang2023vicuna}, and LLaMA \cite{touvron2023llama}, has equipped embodied agents with stronger capabilities in language understanding, task planning, and instruction following. RL offers the core algorithmic framework for agents to learn through interaction with their environments. Representative approaches include DQN \cite{mnih2013playing}, AlphaGo \cite{silver2016mastering}, PPO \cite{schulman2017proximal}, SAC \cite{haarnoja2018soft}, RLHF \cite{ouyang2022training}, and GRPO \cite{shao2024deepseekmath}.

Beyond these classical fields, one of the most promising directions in embodied AI lies in the integration of LLMs/MLLMs with WMs. LLMs and MLLMs (like Flamingo \cite{alayrac2022flamingo}, Qwen-VL \cite{bai2023qwen}, Gemini-1.5 \cite{team2024gemini}, GPT-4o \cite{hurst2024gpt}, and Deepseek-R1 \cite{guo2025deepseek}) provide agents with the ability to understand instructions, reason over multimodal inputs, and generalize across tasks and environments. In contrast, WMs (like Mental Model \cite{ha2018world}, RSSM \cite{hafner2019learning}, JEPA \cite{lecun2022path}, Dreamer-v3 \cite{hafner2023mastering}, Sora \cite{zhu2024sora}, and Genie \cite{bruce2024genie}) enable agents to model and predict environmental dynamics, supporting imagination-based planning and anticipatory decision-making in dynamic and uncertain environments.

\subsubsection{Key Components of Embodied AI} 
Driven by advances in these key technologies, embodied AI has experienced rapid progress. In the following, we present a structured overview of developments in three key components.

\paragraph{Active Perception}
Active perception refers to the agent selectively acquiring information from environmental observations \cite{chaplot2021seal, ahmed2023active, wang2024actiview}. Existing active perception methods can be roughly divided into three categories: visual SLAM, 3D scene understanding, and active environment exploration. To offer an effective perspective on active perception approaches, as summarized in Table \ref{tab:active_perception}, we analyze representative methods along three practical dimensions: sensor type, feature type, and applicable scenarios.

\textbf{Visual SLAM.} Simultaneous Localization and Mapping (SLAM) is a pivotal technology enabling agents to both localize themselves and construct environmental maps in unknown environments \cite{saputra2018visual, macario2022comprehensive}. As a foundational technology of active perception, visual SLAM has been extensively studied \cite{durrant2006simultaneous, bailey2006simultaneous}. According to Wang et al. \cite{wang2024survey}, existing methods fall into geometric-based and semantic-based categories. Geometric methods exploit spatial or temporal cues \cite{mur2015orb}, such as dense scene flow \cite{alcantarilla2012combining, kim2016effective}, triangulation consistency \cite{zou2012coslam}, and graph structure \cite{dai2020rgb, du2020accurate}, performing well in static settings but struggling with dynamic scenes. In contrast, semantic methods improve localization and mapping in dynamic environments by leveraging high-level information. Representative early methods include SLAM++ \cite{salas2013slam++}, integrating object-level semantics, and DS-SLAM \cite{yu2018ds}, applying deep learning to dynamic scene understanding. Recent models such as TwistSLAM \cite{gonzalez2022twistslam} and GS-SLAM \cite{yan2024gs} further enhance robustness by combining geometric optimization with semantic or generative modeling. 

\begin{table*}[t]
\centering
\caption{Comparison of three categories of active perception methods including Visual SLAM, 3D scene understanding, and active environment exploration.}
\label{tab:active_perception}
\renewcommand{\arraystretch}{1.4}
\resizebox{1.\textwidth}{!}{
\begin{tabular}{@{}ccccccc@{}}
\toprule
\textbf{Category} &  \textbf{Method} & \textbf{Year} & \textbf{Sensor Type} & \textbf{Feature Type} & \textbf{Applicable Scenarios} \\
\midrule
\multicolumn{1}{c}{\multirow{6}{*}{Visual SLAM}}
& CoSLAM \cite{zou2012coslam}            & 2012 & RGB‑D             & Geometric + Volumetric    & Dynamic SLAM \\ 
& SLAM++ \cite{salas2013slam++}          & 2013 & RGB‑D             & Semantic                  & Object-level Mapping \\ 
& ORB-SLAM \cite{mur2015orb}             & 2015 & RGB‑D + Stereo    & Geometric                 & Dynamic SLAM \\ 
& DS-SLAM \cite{yu2018ds}                & 2018 & RGB‑D             & Geometric + Semantic      & Dynamic SLAM \\ 
& TwistSLAM \cite{gonzalez2022twistslam} & 2022 & RGB‑D + Stereo    & Geometric + Semantic      & Dynamic SLAM \\ 
& GS-SLAM \cite{yan2024gs}               & 2024 & RGB‑D             & Volumetric                & Object-level Mapping \\ 

\midrule
\multicolumn{1}{c}{\multirow{6}{*}{\makecell[c]{3D Scene\\Understanding}}}
& Gaudi \cite{bautista2022gaudi}          & 2022 & RGB                  & Volumetric    & General Scene Understanding \\   % Immersive
& Clip2Scene \cite{chen2023towards}       & 2023 & RGB + Point Cloud    & Multimodal   & Language-guided Scene Understanding \\
& OpenScene \cite{peng2023openscene}      & 2023 & RGB + Point Cloud    & Multimodal    & General Scene Understanding \\
& Lexicon3D \cite{man2024lexicon3d}       & 2024 & RGB‑D                & Semantic      & Language-guided Scene Understanding \\
& GraphDreamer \cite{gao2024graphdreamer} & 2024 & RGB                  & Topological + Semantic   & Structured Scene Reasoning \\
& HUGS \cite{zhou2024hugs}                & 2024 & RGB‑D                & Multimodal   & General Scene Understanding \\
& RegionPLC \cite{yang2024regionplc}      & 2024 & RGB + Point Cloud & Multimodal   & Language-guided Scene Understanding \\

\midrule
\multicolumn{1}{c}{\multirow{6}{*}{\makecell[c]{Active \\Environment \\Exploration}}}
& MAX \cite{shyam2019model}                     & 2019 & RGB            & Semantic      & Semantic-guided Exploration \\
& Active Neural SLAM \cite{chaplot2020learning} & 2020 & RGB‑D          & Volumetric    & Geometry-based Exploration \\
& APT \cite{liu2021behavior}                    & 2021 & RGB            & Semantic      & Semantic-guided Exploration \\
& Conan \cite{xu2023active}                     & 2023 & RGB            & Topological   & Geometry-based Exploration \\
& DBMF-BPI \cite{russo2023model}                & 2023 & RGB‑D          & Volumetric    & Geometry-based Exploration \\
& ActiveRIR \cite{somayazulu2024activerir}      & 2024 & RGB + Audio    & Multimodal   & Cross-modal Active Perception \\
\bottomrule

\end{tabular}
}
\end{table*}

\begin{table*}[ht]
\centering
\caption{Comparison of three categories of embodied cognition methods: task‑driven self‑planning, memory‑driven self‑reflection, and embodied multimodal foundation models.  \texttt{I}, \texttt{L} and \texttt{P} indicate the Image, Language and Point cloud modalities, respectively.}
\renewcommand{\arraystretch}{1.4}
\resizebox{1.\textwidth}{!}{
\begin{tabular}{@{}cccccccc@{}}
\toprule
\textbf{Category} & \textbf{Method} & \textbf{Year} & \textbf{Input Modalities} & \textbf{Cognition Type} & \textbf{Reasoning Mode} & \textbf{Output} \\
\midrule
\multicolumn{1}{c}{\multirow{6}{*}{\makecell[c]{Task-driven \\Self-planning}}}
& L3P \cite{zhang2021world}         & 2021 & I + L  & Planner   & Neural + Symbolic          & Action \\
& LLM‑Planner \cite{song2023llm}    & 2023 & I + L  & Planner   & Neural + Symbolic & Action \\
& Egoplaner \cite{zhou2020ego}      & 2023 & I      & Planner   & Symbolic          & Action \\
& AutoAct \cite{qiao2024autoact}    & 2024 & L      & Planner   & Neural            & Action \\
& RPG \cite{yang2024mastering}      & 2024 & I + L  & Planner   & Neural            & Policy \\
& ETPNav \cite{an2024etpnav}        & 2024 & I + L  & Planner   & Neural + Symbolic & Policy \\
\midrule
\multicolumn{1}{c}{\multirow{6}{*}{\makecell[c]{Memory‑driven \\Self‑reflection}}}
& Reflexion \cite{shinn2023reflexion}   & 2023 & L      & Memory    & Beam + Replay     & Policy \\
& Reflect \cite{liu2023reflect}         & 2023 & I + L  & Memory    & Neural + Symbolic & Policy \\
& RILA \cite{yang2024rila}              & 2024 & L      & Memory    & Neural            & Policy \\
& Optimus‑1 \cite{li2024optimus}        & 2024 & I + L  & Memory    & Neural            & Policy \\
& EvoAgent \cite{feng2025evoagent}      & 2025 & I + L  & Memory    & Neural            & Policy \\
& REMAC \cite{yuan2025remac}            & 2025 & L      & Memory    & Neural + Symbolic & Policy \\
\midrule
\multicolumn{1}{c}{\multirow{6}{*}{\makecell[c]{Embodied Multimodal \\Foundation Models}}}
& SayCan \cite{brohan2023can}         & 2022 & I + L        & Planner + Aligner & Neural    & Answer  + Action  \\
& GATO \cite{reed2022generalist}       & 2022 & I + L + P   & Aligner           & Neural    & Action            \\
& EmbodiedGPT \cite{mu2023embodiedgpt}& 2023 & I + L        & Aligner           & Neural    & Answer  + Action  \\
& Kosmos‑2 \cite{peng2023kosmos}       & 2023 & I + L       & Aligner           & Neural    & Answer            \\
& MultiPLY \cite{hong2024multiply}     & 2024 & I + L       & Aligner           & Neural    & Answer            \\
& ManipLLM \cite{li2024manipllm}       & 2024 & I + L       & Aligner           & Neural    & Answer + Action   \\
\bottomrule
\end{tabular}
}

\label{tab:embodied_cognition}
\end{table*}

\begin{table*}[ht]
\centering
\caption{Comparison of three categories of dynamic interaction methods including action control, behavioral interaction, and collaborative decision-making, across input modalities, interaction type, modeling paradigm, and task type. \texttt{I}, \texttt{L}, \texttt{S}, \texttt{P}, and \texttt{T} denote Image, Language, State, Proprioception, and Trajectory, respectively. IL denotes Imitation Learning.}
\renewcommand{\arraystretch}{1.4}
\resizebox{1.\textwidth}{!}{
\begin{tabular}{@{}cccccccc@{}}
\toprule
\textbf{Category} & \textbf{Method} & \textbf{Year} & \textbf{Input Modalities} & \textbf{Interaction Type} & \textbf{Learning Paradigm} & \textbf{Task Type} \\
\midrule
\multicolumn{1}{c}{\multirow{8}{*}{\makecell[c]{Action \\ Control}}}
& MineDojo \cite{fan2022minedojo}       & 2022 & I + L     & High-level Planning & LLM   & Instruction Following \\
& PaLM-E \cite{driess2023palm}          & 2023 & I + L + P & Low-level Control   & MLLM  & Embodied Manipulation \\
& RT-2 \cite{brohan2023rt}              & 2023 & I + L     & Low-level Control   & VLA   & Embodied Manipulation \\
& OpenVLA \cite{kim2024openvla}         & 2024 & I + L     & Low-level Control   & VLA   & Embodied Manipulation \\
& Cogagent \cite{hong2024cogagent}      & 2024 & I + L     & Low-level Control   & MLLM  & Instruction Following \\
& Octo \cite{team2024octo}              & 2024 & I + L + P & Low-level Control   & VLA   & Embodied Manipulation \\
& CrossFormer \cite{doshi2024scaling}   & 2024 & I + L     & Low-level Control   & VLA   & Embodied Manipulation \\
& HPT \cite{wang2024scaling}            & 2024 & I + L     & Low-level Control   & VLA   & Embodied Manipulation \\
\midrule
\multicolumn{1}{c}{\multirow{7}{*}{\makecell[c]{Behavioral \\ Interaction}}}
& GAIL \cite{ho2016generative}          & 2016 & T     & Behavioral             & IL & Trajectory Learning \\
& MGAIL \cite{baram2017end}             & 2017 & T     & Behavioral             & IL & Trajectory Learning \\
& TrafficSim \cite{suo2021trafficsim}   & 2021 & T     & Behavioral             & RL & Trajectory Learning \\
& TrajGen \cite{zhang2022trajgen}       & 2022 & I + T & Behavioral             & RL & Trajectory Learning \\
& Behavior-1K \cite{li2023behavior}     & 2023 & I     & Trajectory             & IL & Behavior Understanding \\
& AgentLens \cite{lu2024agentlens}      & 2024 & I + S & Trajectory             & IL & Behavior Understanding \\
& ECL \cite{chen2024embodied}           & 2024 & I + L & High-level Planning    & IL & Embodied Manipulation \\
\midrule
\multicolumn{1}{c}{\multirow{8}{*}{\makecell[c]{Collaborative \\ Decision}}}
& QMIX \cite{rashid2020monotonic}       & 2018 & S     & Behavioral   & RL  & Cooperative Decision \\
& Qtran \cite{son2019qtran}             & 2019 & S     & Behavioral   & RL  & Cooperative Decision \\
& QPLEX \cite{wang2020qplex}            & 2019 & S     & Behavioral   & RL  & Cooperative Decision \\
& MAT \cite{wen2022multi}               & 2022 & S     & Behavioral    & RL  & Cooperative Decision \\
& CoELA \cite{zhang2023building}        & 2024 & I + L & Low-level Control    & LLM & Cooperative Manipulation \\
& AgentVerse \cite{chen2023agentverse}  & 2024 & L     & High-level Planning    & LLM & Agent Society Simulation \\
& MetaGPT \cite{hong2023metagpt}        & 2024 & L     & High-level Planning    & LLM & Agent Society Simulation \\
& Combo \cite{zhang2024combo}           & 2024 & L     & High-level Planning    & LLM & Cooperative Planning \\
\bottomrule
\end{tabular}}
\label{tab:dynamic_interaction}
\end{table*}

\textbf{3D Scene Understanding.} Scene understanding focuses on enabling agents to perceive, segment, and reason about complex environments in a structured and semantically meaningful way. Recent works have advanced this field by integrating vision-language models and generative priors. Early efforts like Gaudi \cite{bautista2022gaudi} introduced generative models for 3D-aware scene synthesis. Clip2Scene \cite{chen2023towards} and OpenScene \cite{peng2023openscene} leveraged vision-language embeddings to facilitate label-efficient and open-vocabulary 3D understanding. Structured scene understanding is further enhanced by Lexicon3D \cite{man2024lexicon3d} and GraphDreamer \cite{gao2024graphdreamer}, which model object-level relations in 3D space through structured representations such as scene graphs or semantic lexicons.  
Meanwhile, region-level multimodal grounding techniques, exemplified by HUGS \cite{zhou2024hugs} and RegionPLC \cite{yang2024regionplc}, incorporate prompts and spatial grounding to achieve fine-grained, goal-conditioned 3D perception. These methods advance holistic, language-aligned 3D understanding.

\textbf{Active Environment Exploration.} Active exploration focuses on enabling agents to autonomously acquire informative observations through interaction with the environment. Early approaches relied on building explicit or implicit environmental models. Representative model-based methods include MAX \cite{shyam2019model} and Active Neural SLAM \cite{chaplot2020learning}, which leverage predictive modeling and mapping to support efficient navigation in unseen spaces. In contrast, APT \cite{liu2021behavior} and DBMF-BPI \cite{russo2023model} focus on model-free exploration through direct environmental interaction to reduce reliance on explicit modeling. Recent efforts further enhance exploration capabilities by incorporating multimodal perception \cite{somayazulu2024activerir} and semantic reasoning \cite{xu2023active}. 

\paragraph{Embodied Cognition}
Embodied cognition refers to the emergence of internal representations and reasoning capabilities during the interaction, driven by the agent's self-reflection on its perception and accumulated experience \cite{clark1999embodied, gibbs2005embodiment, ziemke2013s}. This component forms the core of embodied AI, enabling agents to perform task planning \cite{liu2023egocentric}, causal inference \cite{stocking2022robot}, and long-horizon reasoning \cite{sermanet2024robovqa, zhang2024fltrnn}. Recent studies of embodied cognition primarily focus on three aspects: task-driven self-planning, memory-driven self-reflection, and embodied multimodal foundation models. Table~\ref{tab:embodied_cognition} presents representative methods analyzed from four perspectives: input modalities, cognition type, reasoning mode, and output type. These dimensions reflect how embodied agents perceive information, form internal models and conduct reasoning.

\textbf{Task-driven Self-Planning.} In task-driven self-planning, agents autonomously generate structured plans based on task goals, environmental context, and internal knowledge, without explicit human instructions \cite{nayak2024long, zhang2024lamma, padmakumar2022teach}. Structured learning is a classical solution that develops latent planning spaces or direct policy mappings, achieving high efficiency within training distributions but lacking robustness to out-of-distribution scenarios. Representative approaches include L3P \cite{zhang2021world}, Egoplaner \cite{zhou2020ego}, and ETPNav \cite{an2024etpnav}. Recent advances incorporate LLMs or generative models into self-planning. LLM-Planner \cite{song2023llm} and AutoAct \cite{qiao2024autoact} integrate LLMs into planning by grounding language-guided reasoning into various tasks, while RPG \cite{yang2024mastering} offers a generative perspective, aiming to unify planning and content creation through multimodal reasoning.

\textbf{Memory-driven Self-Reflection.} Memory-driven self-reflection enables agents to leverage past experiences for long-horizon reasoning, error correction, and self-improvement \cite{zhang2024survey, feng2025evoagent}. Early studies focus on memory processing, including fixed-size replay buffers \cite{bakker2001reinforcement, ramani2019short, zhu2020episodic} and differentiable memory architectures \cite{khan2017memory, pritzel2017neural}. Recent advances introduce reflective mechanisms, where agents summarize or verbalize past experiences to guide future decisions. Reflexion \cite{shinn2023reflexion} and Reflect \cite{liu2023reflect} enable agents to iteratively self-correct by integrating verbalized feedback into action planning, while RILA \cite{yang2024rila} extends reflective reasoning to multimodal semantic navigation. Beyond individual reflection, Optimus-1 \cite{li2024optimus} and REMAC \cite{yuan2025remac} integrate multimodal or multi-agent memory to support long-horizon collaboration. EvoAgent \cite{feng2025evoagent} further advances this direction by coupling continual world modeling with a memory-driven planner, enabling fully autonomous evolution across sequential tasks.

\textbf{Embodied Multimodal Foundation Models.} In the era of MLLMs, embodied multimodal foundation models \cite{xu2024survey, ren2024embodied, firoozi2023foundation} have emerged as one of the most promising solutions for unifying planning, reasoning, and other embodied cognitive capabilities. Recent progress is driven by both data construction and model development. Data efforts focus on constructing high-quality benchmarks to support scalable and cognitively meaningful evaluation, such as MuEP \cite{li2024muep}, ECBench \cite{dang2025ecbench}, MFE-ETP \cite{zhang2024mfe}, and EmbodiedBench \cite{yang2025embodiedbench}. On the model side, recent advances include affordance-grounded agents (e.g., SayCan \cite{brohan2023can} and GATO \cite{reed2022generalist}) that align language understanding with embodied action spaces, vision-language pretraining approaches (like EmbodiedGPT \cite{mu2023embodiedgpt} and Kosmos-2 \cite{peng2023kosmos}) that promote scalable embodied reasoning, and object-centric designs (such as MultiPLY \cite{hong2024multiply} and ManipLLM \cite{li2024manipllm}) that enhance manipulation and interaction capabilities. These models collectively aim to build transferable and generalizable embodied AI.

\paragraph{Dynamic Interaction} 
Dynamic interaction refers to the process in which an agent influences the environment through actions or behaviors grounded in its perception and cognition \cite{sun2024comprehensive, jin2025embodied}. Existing research highlights the significance of this capability in enabling agents not only to respond but also to change their surroundings \cite{billard2019trends, ajoudani2018progress}. Studies on dynamic interaction encompass action control, behavioral interaction, and collaborative decision-making. To better understand existing methods, we analyze representative approaches from four perspectives, including input modalities, interaction type, learning paradigm, and task type, as shown in Table \ref{tab:dynamic_interaction}. These dimensions reflect how agents sense the environment, determine the level and structure of interaction, and generate appropriate behaviors in dynamic multi-agent or human-in-the-loop scenarios.

\textbf{Action Control.} Action control generates motor commands for embodied interaction. Early methods were based on control theory with dynamic system modeling \cite{kelly1995tuning, rocco2002stability} or RL via trial and error \cite{kiumarsi2017optimal, singh2022reinforcement}. The former is effective for structured or repetitive tasks, while the latter is adaptable to high-dimensional, nonlinear problems. Recent advances mainly follow three directions. Vision-language-action (VLA) models, such as PaLM-E \cite{driess2023palm}, RT-2 \cite{brohan2023rt}, OpenVLA \cite{kim2024openvla}, and CogAgent \cite{hong2024cogagent}, integrate language-guided reasoning for flexible control and have been comprehensively reviewed by Ma et al. \cite{ma2024survey}. Open-ended frameworks like MineDojo \cite{fan2022minedojo} promote continual skill acquisition from open-world knowledge. In addition, Cross-embodiment learning, including CrossFormer \cite{doshi2024scaling}, HPT \cite{wang2024scaling}, and Octo \cite{team2024octo}, aim to unify policy learning across diverse robots and modalities. 

\textbf{Behavioral Interaction.} The behavior of an agent is composed of a sequence of actions. Compared to action control, it emphasizes high-level control through meaningful action patterns, enabling agents to interact in a flexible and goal-directed manner. Recent advances mainly fall into two directions. Imitation learning, including GAIL \cite{ho2016generative}, MGAIL \cite{baram2017end}, TrafficSim \cite{suo2021trafficsim}, and TrajGen \cite{zhang2022trajgen}, enables efficient acquisition and simulation of complex behaviors. BEHAVIOR-1K \cite{li2023behavior} provides a large-scale benchmark for evaluating behavior generalization across 1,000 embodied tasks. Behavior-aware enhancement methods, such as AgentLens \cite{lu2024agentlens} and ECL \cite{chen2024embodied}, improve policy robustness and interpretability. Despite these advances, achieving reliable long-horizon behavioral interaction under sparse feedback remains challenging.

\textbf{Collaborative Decision.} Collaborative decision focuses on coordinating multiple agents to achieve shared goals, which is essential for multi-agent systems and human-robot collaboration \cite{jiang2024multi, zhu2024madiff, ma2024human}. Multi-agent RL is a classical solution, with methods like QTRAN \cite{son2019qtran}, QPLEX \cite{wang2020qplex}, and Qatten \cite{rashid2020monotonic} addressing cooperation via centralized training with decentralized execution. MAT \cite{wen2022multi} reframes MARL as a sequence modeling problem to mitigate scalability limitations in multi-agent RL. Recent advances integrate LLMs and WMs to enhance multi-agent collaboration. MetaGPT \cite{hong2023metagpt}, CoELA \cite{zhang2023building}, and AgentVerse \cite{chen2023agentverse} leverage LLMs for task reasoning and coordination, while COMBO \cite{zhang2024combo} composes modular WMs to support scalable collaborative embodied decision.

\vspace{-0.5em}
\subsection{Hardware} \label{hardware} 
As embodied AI evolves, model complexity and size have grown, increasing computational and energy demands. Embodied systems, often operating in dynamic, real-world environments, face strict latency and power constraints—especially at the edge. Thus, developing hardware-friendly directions that maintain performance while optimizing efficiency is crucial for enabling responsive, energy-aware embodied agents. Hardware optimization in embodied AI typically includes four components: hardware-aware model compression, compiler-level optimization, domain-specific accelerators, and hardware-software co-design.

\subsubsection{Hardware-aware Model Compression}
Quantization and pruning~\cite{han2015deep} are key techniques for reducing model size and computational cost. In embodied agents, which frequently run on low-power embedded hardware, such techniques are vital for enabling fast and efficient inference. Quantization~\cite{xiao2023smoothquant} maps weights and activations to lower bit-widths, while pruning~\cite{wang2021spatten} removes redundant parameters. To support real-world embodied tasks, such as robotic control or visual navigation, hardware efficiency metrics like power, performance, and area (PPA) can guide bit-width allocation or pruning ratios~\cite{wang2019haq}, enabling task-specific trade-offs between accuracy and deployability on physical platforms.

\subsubsection{Compiler-level Optimization}
Compilers bridge high-level embodied AI models and hardware execution. In real-time embodied systems, compiler toolchains are essential for efficient processing of sensor data and decision-making. TVM~\cite{chen2018tvm}, built on LLVM~\cite{lattner2004llvm} and CUDA, generates optimized code across platforms. These compilers transform computational graphs through operator fusion and redundant computation elimination~\cite{sun2022gtuner}, enabling responsive behavior. Mapping strategies like loop reordering and tiling enhance data locality, parallelism, and memory access~\cite{cai2024gemini}, all of which are critical to maintaining low-latency inference in embodied agents.

\subsubsection{Domain-specific Accelerators}
With growing computational demands, domain-specific accelerators (DSAs) are a promising solution for embodied AI. These systems, from robots to AR/VR agents, benefit from fast, energy-efficient hardware tailored for frequent operations. Google’s TPU~\cite{jouppi2017datacenter}, typically integrated with CPUs and GPUs via PCIe, accelerates key operations like matrix multiplication. FPGA-based accelerators~\cite{liu2025flightvgm} allow reconfigurability for adapting to new tasks or changing workloads; CGRA accelerators~\cite{qin2025picachu} improve structured, dataflow-heavy computations common in perception or control. Meanwhile, ASIC-based accelerators~\cite{genc2021gemmini} offer high throughput and energy efficiency, ideal for deploying high-performance embodied models in real-world environments.

\subsubsection{Hardware-software Co-design}
Separating algorithm and hardware design can lower runtime efficiency. Hardware-software co-design addresses this through algorithm-system and algorithm-hardware co-optimization. Algorithm-system co-optimization focuses on how to take full advantage of GPU resources like tensor cores and CUDA cores to better support the algorithm~\cite{lin2024qserve}. Algorithm-hardware co-optimization aims to improve deployment efficiency by tuning both the model and the hardware architecture. For example, we can perform multi-objective optimization based on the types of operators in the network and the configuration parameters of the hardware~\cite{wang2025jaq}. We can also design different numerical quantization schemes along with matching hardware accelerators to better support embodied AI tasks~\cite{zou2024bie}.

\subsection{Benchmarks and Evaluation Metrics}
Standardized benchmarks and evaluation metrics are crucial for objectively assessing the performance of embodied AI systems. Widely adopted testbeds include Habitat \cite{savva2019habitat}, which provides photorealistic 3D indoor environments for navigation and interaction tasks, and ManiSkill \cite{mu2021maniskill}, offering physics-based manipulation scenarios with diverse object sets. Simulation platforms like MuJoCo \cite{todorov2012mujoco} enable precise control evaluation in continuous state-spaces, while EmbodiedBench \cite{yang2025embodiedbench} supports holistic evaluation of vision-driven agents across perception, cognition, and interaction. For UAV applications, AirSim \cite{shah2017airsim}, U2UData \cite{feng2024u2udata} and U2USim \cite{u2usim} provides high-fidelity aerial environments with dynamic obstacles. These testbeds vary in complexity: Habitat excels in visual realism, ManiSkill in object diversity, MuJoCo in physical accuracy, and EmbodiedBench in multimodal integration. Domain-specific benchmarks like BEHAVIOR-1K \cite{li2023behavior} further enable granular evaluation of 1,000 everyday activities under realistic constraints.

Key evaluation metrics span three critical dimensions: Task Success Rate measures completion accuracy of goal-oriented objectives (e.g., object manipulation or navigation) \cite{brohan2023rt}; Real-time Responsiveness quantifies decision latency and adaptation speed to environmental changes \cite{driess2023vision}; and Energy Efficiency evaluates computational cost (FLOPS) and power consumption (Watts) during deployment \cite{han2015deep}. Additional metrics include Path Length for navigation efficiency \cite{chaplot2020learning}, Generalization Score for unseen scenarios \cite{openvla2024}, and Safety Violations for physical compliance \cite{billard2019trends}. For multi-agent systems, Coordination Efficiency \cite{jiang2024multi} and Communication Overhead \cite{zhang2024towards} provide critical insights. Standardized evaluation protocols like those in MFE-ETP \cite{zhang2024mfe} ensure fair cross-modal comparisons, though challenges remain in sim-to-real transfer validation \cite{ma2024survey}.

\begin{figure}[t]
\centering
\includegraphics[width=\linewidth]{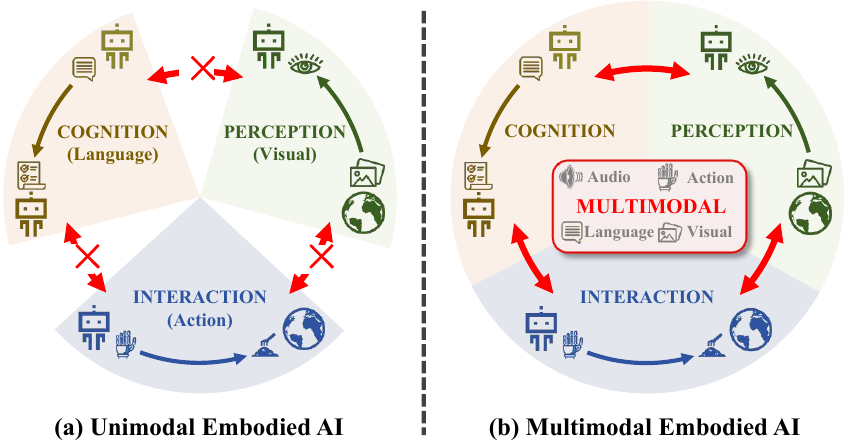}
\caption{Unimodal embodied AI and multimodal embodied AI. (a) Unimodal methods focus on specific modules of embodied AI. They are limited by the narrow scope of information provided by each modality and the inherent gaps between modalities across modules. (b) Multimodal embodied AI methods break these limitations and enable the mutual enhancement of the modules.}
\label{fig:modal}
\vspace{-1em}
\end{figure}

\vspace{-0.8em}
\subsection{From Unimodal to Multimodal} \label{development}
The development of embodied AI has evolved from unimodal to multimodal systems, as shown in Fig \ref{fig:modal}. Initially, embodied AI was primarily concerned with individual modalities, such as vision, language, or action, where the perception, cognition, and interaction were driven by one sensory input \cite{engel2014lsd, mur2015orb}. As the field matured, the limitations of unimodal embodied AI became apparent, and there has been a significant shift toward integrating multiple sensory modalities  \cite{driess2023palm, jiang2024fisher, wang2024actiview}. Multimodal embodied AI is now seen as crucial for creating more adaptive, flexible, and robust agents capable of performing complex tasks in dynamic environments \cite{qin2024mp5, yang2025embodiedbench}.

Unimodal embodied AI has benefited from rapid developments in fundamental areas such as computer vision, natural language processing, and reinforcement learning \cite{jayaraman2018learning, lin2023grounded}. These unimodal methods excel in dealing with a specific module in embodied AI. For example, computer vision techniques have driven advances in visual SLAM and 3D scene understanding in the active perception module \cite{macario2022comprehensive}. Natural language processing techniques, especially LLM, have become popular solutions to address task planning and long-horizon reasoning in the embodied cognition module \cite{liu2023embodied, lu2023llm}. Although unimodal embodied AI performs well in independent modules, it always faces two inherent limitations. On the one hand, the information contained in a single modality is limited, hindering the performance of perception, cognition, and interaction. For example, visual-only systems struggle to understand environments in dynamic or ambiguous settings, while auditory-based systems face challenges in real-world noise and signal processing \cite{wang2024embodiedscan, qin2024mp5}. On the other hand, diverse and heterogeneous modalities hinder information transfer and sharing among modules. The agent's perception of the environment fails to facilitate the formation of its cognition, while the evolution of cognition fails to facilitate the interaction with the environment.

In contrast, multimodal embodied AI has emerged as a more promising paradigm \cite{yang2025embodiedbench}. By integrating data from multiple sensing modalities, such as visual, auditory, and olfactory feedback, these methods can provide a more holistic and precise understanding of the environment. More importantly, multimodal embodied AI can facilitate deeper integration among perception, cognition, and interaction. Recent advances in MLLMs and WMs enable agents to more effectively handle multiple modalities, promising to improve the capabilities of embodied AI \cite{wang2024drivedreamer, wu2023daydreamer, zhu2024sora}. The integration of these models is considered a key step toward enabling multimodal embodied AI in dynamic, uncertain environments.
\section{Embodied AI with LLMs/MLLMs}

\begin{figure*}[t]
\centering
\includegraphics[width=0.82\textwidth]{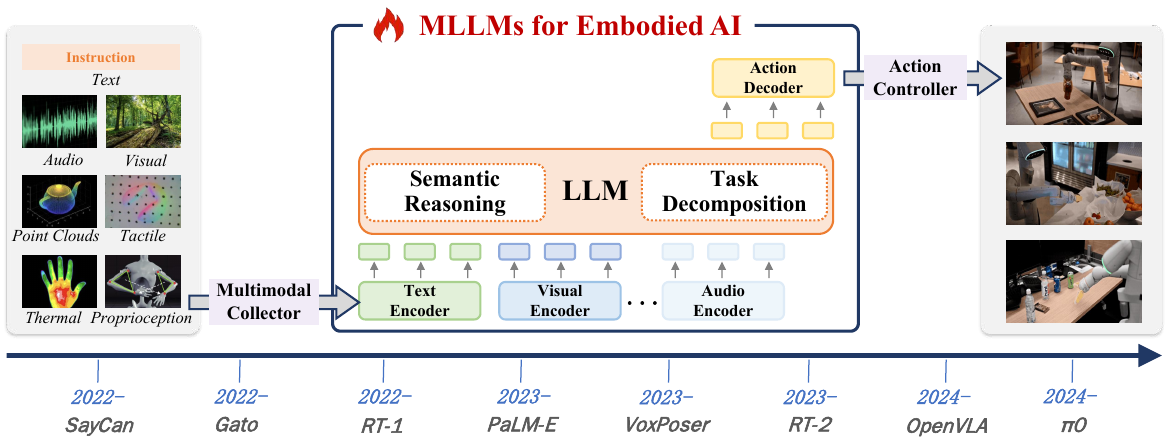}
\caption{The development roadmap of MLLMs for embodied AI. This roadmap highlights the key milestones in their conceptual and practical development.}
\label{fig:3_1}
\vspace{-1em}
\end{figure*}

This section provides a comprehensive overview of embodied AI with LLMs/MLLMs. We first elaborate in detail how LLMs boost embodied AI in Subsection \ref{HOW_LLMS} and how MLLMs boost embodied AI in Subsection \ref{HOW_MLLMS}. Then we discuss the classification of MLLMs for embodied AI in Subsection \ref{CLASS_MLLMS}.

\vspace{-0.7em}
\subsection{LLMs Boost Embodied AI}
\label{HOW_LLMS}
LLMs empower embodied AI via semantic reasoning and task decomposition, bringing high-level natural language instructions and low-level natural language actions into embodied cognition. 

\subsubsection{Semantic Reasoning} Semantic reasoning \cite{brohan2022rt, zhao2025embodied, songmodularized} leverages LLMs to interpret semantics from text instructions by analyzing linguistic patterns \cite{ge2025dynamic}, contextual relationships \cite{huang2024neighbor}, and implicit knowledge \cite{zhang2024large}. Through transformer architectures \cite{vaswani2017attention}, LLMs map input tokens to latent representations, enabling hierarchical abstraction of meaning across syntactic and pragmatic levels. They employ attention mechanisms to weigh relevant semantic cues while suppressing noise, facilitating logical inference and analogical reasoning. By integrating world knowledge from pretraining corpora with task-specific prompts, LLMs dynamically construct conceptual graphs that align textual inputs with intended outcomes. This process supports multi-hop reasoning through probabilistic token prediction, resolving ambiguities by evaluating contextual coherence and semantic plausibility. 

\subsubsection{Task Decomposition} Task decomposition \cite{alayrac2022flamingo, floridi2020gpt} employs LLMs’ sequential logic to break complex objectives into sub-tasks by hierarchically analyzing contextual dependencies and goal alignment. Leveraging chain-of-thought prompting, LLMs iteratively parse instructions into actionable steps, prioritizing interdependencies while resolving ambiguities through semantic coherence checks. 

Representative works like SayCan \cite{ahn2022saycan} first provides a real-world pretrained natural language actions library, which is used to constrain LLMs to propose both feasible and contextually appropriate actions; then uses LLMs to convert natural language instructions into natural language action sequences; finally uses value functions to verify the feasibility of natural language action sequences in a particular physical environment. These works suggest that LLMs are extremely useful to robots aiming to act upon high-level, temporally extended instructions expressed in natural language. However, LLMs are only a part of the entire embodied AI system, which is limited by a fixed natural language actions library and a specific physical environment, and it is difficult to achieve adaptive expansion in new robots and environments.

\vspace{-0.5em}
\subsection{MLLMs Boost Embodied AI}
\label{HOW_MLLMS}
MLLMs can act on the entire embodied AI system and can solve LLMs' problems well by bridging high-level multimodal inputting \cite{pan2025modular} and low-level motor action sequences \cite{huang2024vtimellm} into end-to-end embodied applications (as shown in Fig. \ref{fig:3_1}). Compared with LLMs, semantic reasoning \cite{li2024manipllm, duan2022survey, brophy2023review} leverages MLLMs’ cross-modal comprehension to interpret semantics from visual, auditory, or tactile inputs, e.g., identifying objects, inferring spatial relationships, or predicting environmental dynamics. Concurrently, task decomposition \cite{black2410pi0, huang2024rekep, zhao2025cot} employs MLLMs’ sequential logic to break complex objectives into sub-tasks while dynamically adapting plans based on sensor feedback. MLLMs mainly include Vision-Language Models (VLMs) and Vision-Language-Action models (VLAs).

\subsubsection{VLMs for Embodied AI}
VLMs for embodied AI integrate visual and language instruction understanding to enable physical or virtual agents to perceive their environments in goal-driven tasks \cite{wang2023curriculum, wang2025automated, wang2022disentangled}. Representative works like PaLM-E \cite{driess2023palm} first train visual and language encodings end-to-end, in conjunction with a pre-trained large language model; then incorporate the results of real-world continuous sensor modalities encodings into VLMs and establish the link between words and percepts; finally, achieve multi-task completion through fixed action space mapping. For navigation, ShapeNet \cite{thomason2022language}, which fine-tunes contrastive embeddings for 3D spatial reasoning, greatly reduces path planning errors. These works suggest that VLMs can combine perception and reasoning in embodied AI to solve a large number of tasks with fixed action spaces.

\subsubsection{VLAs for Embodied AI}
VLAs integrate multimodal inputs with low-level action control through differentiable pipelines. Representative works like RT-2 \cite{brohan2023rt} first encode the robot's current image, language instructions, and robot actions at a specific timestep and convert them into text tokens; then use LLMs for semantic reasoning and task decomposition; finally, de-tokenizes generated tokens into the final action. Octo \cite{team2024octo} pretrains on 100K robot demonstrations with language annotations, achieving cross-embodiment tool use. For dexterous manipulation, PerAct \cite{peract2023} utilizes 3D voxel representations to reach millimeter-level grasp accuracy. These works suggest that VLAs can act on the entire embodied AI system and achieve adaptive expansion in new robots and environments.

\vspace{-0.5em}
\subsection{Classification of MLLMs for Embodied AI}
\label{CLASS_MLLMS}
MLLMs can empower active perception, embodied cognition, and dynamic interaction of embodied AI.

\subsubsection{MLLMs for Active Perception}
First, MLLMs can enhance 3D SLAM. By grounding visual observations into semantic representations, MLLMs augment traditional SLAM pipelines with high-level contextual information such as object categories, spatial relations, and scene semantics \cite{chen2023curriculum, wang2025divico}. Representative works like SEO-SLAM~\cite{hong2024seoslam} utilize MLLMs to generate more specific and descriptive labels for objects, while dynamically updating a multiclass confusion matrix to mitigate biases in object detection. Second, MLLMs can enhance 3D scene understanding. Camera-based perception~\cite{brophy2023review} remains the dominant setup in MLLM-driven embodied AI, as RGB inputs align naturally with the visual-language pretraining of many foundation models~\cite{yang2025magma, chen2024disenstudio, chen2023disenbooth}. Representative works like EmbodiedGPT~\cite{mu2023embodiedgpt} leverage this synergy to map 2D visual inputs into semantically rich features aligned with language-based goals. Finally, MLLMs can enhance active environment exploration. MLLMs have also revolutionized how robots interact with their environments, particularly in feedback-driven closed-loop interactions. Representative works like LLM$^3$~\cite{wang2024llmˆ} focus on structured motion-level feedback, which incorporates signals such as collision detections into the planning loop, allowing the model to iteratively revise symbolic action sequences. MART~\cite{yue2024mllmretrieval}, on the other hand, leverages interaction feedback to improve retrieval quality.

\subsubsection{MLLMs for Embodied Cognition} First, MLLMs can enhance task-driven self-planning \cite{yao2024data, qin2023multi, lan2023curriculum}. Embodied agents with MLLMs can either directly map high-level goals to structured action sequences~\cite{black2410pi0}, or adopt an intermediate planning strategy that continually interacts with the environment to refine their plans~\cite{huang2024rekep}. Representative works like CoT-VLA~\cite{zhao2025cot} predict intermediate subgoal images that depict the desired outcomes of subtasks, helping the agent visualize and reason through each step of a complex task. Second, MLLMs can enhance memory-driven self-reflecting. MLLMs allow agents to learn from experience using this inherent memory module~\cite{team2024octo}. Representative works like Reflexion~\cite{shinn2023reflexion} enhance agent performance through self-generated linguistic feedback, which is stored in an episodic memory buffer and leveraged to guide future planning. Finally, MLLMs can enhance embodied multimodal foundation models. MLLMs can be adapted to the physical world through continued pretraining or fine-tuning in embodied settings. Representative works include Qwen-VL~\cite{bai2023qwen} and InternVL~\cite{chen2024internvl}, along with models supporting broader modality alignment, such as Qwen2.5-Omni~\cite{xu2025qwen2}.

\begin{figure*}[t]
\centering
\includegraphics[width=0.82\textwidth]{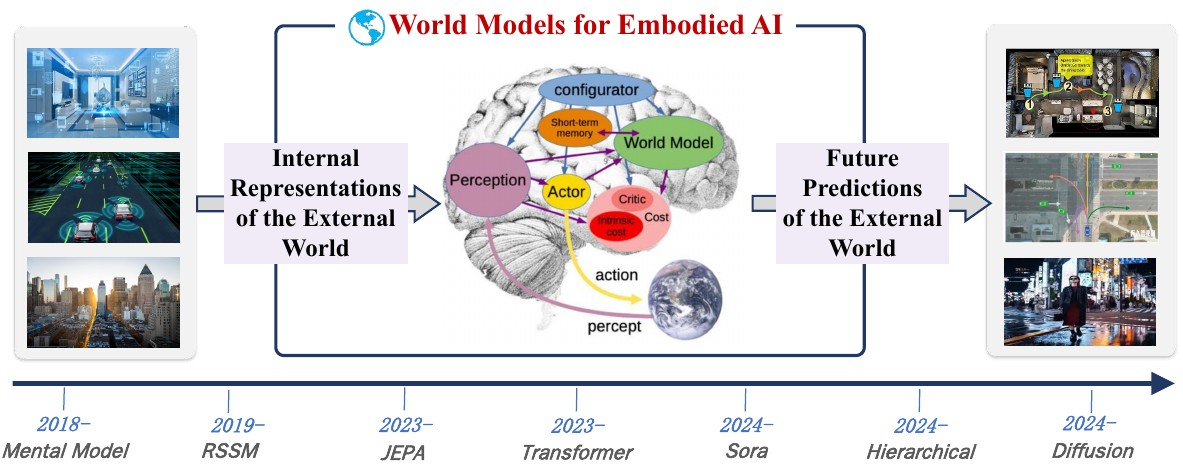}
\caption{The development roadmap of WMs for embodied AI. This roadmap highlights the key milestones in their conceptual and practical development.}
\label{fig:4_1}
\vspace{-1em}
\end{figure*}

\subsubsection{MLLMs for Dynamic Interaction}
First, MLLMs can enhance action control. MLLMs have ability to decompose complex tasks into actionable subtasks~\cite{huang2024rekep}. To further produce continuous control signals for each subtask, MLLMs either generate actions autoregressively in a sequential manner~\cite{zitkovich2023rt, kim2024openvla} or employ auxiliary policy heads to further process their internal representations~\cite{team2024octo}. Recent advances also explore generating executable code with MLLMs~\cite{wang2024executable}, enabling robots to follow interpretable and adaptable control policies. Second, MLLMs can enhance behavioral interaction. Through interaction with the environment, MLLMs are also capable of generating sequences of behavioral actions in a single step. Representative works like $\pi$-0~\cite{black2410pi0} combine a vision-language backbone with a flow-matching decoder to produce smooth, temporally extended behavioral trajectories. Finally, MLLMs can enhance collaborative decision-making. One line of research focuses on multi-agent systems that aim to achieve human-level coordination and adapt rapidly to unforeseen challenges~\cite{wu2025generative}. For instance, Combo~\cite{zhang2024combo} introduces a novel framework that enhances cooperation among decentralized agents operating solely with egocentric visual observations. Other efforts investigate human-agent collaboration. VLAS~\cite{zhao2025vlas} exemplifies this by aligning human verbal commands with visual context via a speech encoder and a LLaVA-style MLLM~\cite{liu2023visual}, enabling fluid and conversational human-agent interaction. 
\section{Embodied AI with World Models}
This section provides a comprehensive overview of embodied AI with WMs. We first elaborate in detail how WMs boost embodied AI in Subsection \ref{HOW_WMS}. Then we discuss the classification of WMs for embodied AI in Subsection \ref{CLASS_WMs}.
 
\vspace{-0.5em}
\subsection{World Models Boost Embodied AI}
\label{HOW_WMS}
WMs empower embodied AI by building internal representations and future predictions of the external world (as shown in Fig. \ref{fig:4_1}), facilitating physical law-compliant embodied interactions in dynamic environments. 

\subsubsection{Internal Representations of the External World}
Internal representations compress rich sensory inputs into structured latent spaces, capturing object dynamics, physics laws, and spatial structures, allowing agents to reason about "what exists" and "how things behave" in their surroundings. These latent embeddings preserve hierarchical relationships \cite{forrester1971counterintuitive} between entities and environments, mirroring the compositional nature of reality itself. The structured nature of these representations facilitates generalization across environments, as abstracted principles (like gravity or object permanence) transcend specific instances. Moreover, they support counterfactual reasoning \cite{wu2023paragraph} by maintaining disentangled variables for objects' intrinsic properties \cite{robine2023transformer} and extrinsic relations  \cite{wang2024worlddreamer}, enabling flexible mental manipulation of individual components. This disentanglement also enhances sample efficiency in learning, as agents transfer knowledge between tasks, sharing latent factors. World models with rich internal representations, can introspect on their own uncertainty about environmental states and actively seek information to resolve ambiguities. By encoding temporal continuity and spatial topology \cite{bruce2024genie}, these models naturally enforce consistency constraints during planning, filtering physically implausible actions before execution. Ultimately, such structured latent spaces act as cognitive scaffolding for building causal understanding \cite{chen2022transdreamer}, mirroring how humans develop intuitive theories about their world through compressed sensory experiences.

\subsubsection{Future Predictions of the External World}
Future predictions simulate potential rewards of sequence actions across multiple time horizons aligned with physical laws, thereby preempting risky or inefficient behaviors \cite{haftner2020dream, haftner2021mastering}. This predictive capacity bridges short-term actions with long-term goals \cite{okada2022dreamingv2}, filtering out trajectories violating physical plausibility (e.g., walking through walls) or strategic coherence (e.g., depleting resources prematurely). Long-horizon prediction \cite{wu2023daydreamer} allows adaptive balancing of exploration-exploitation tradeoffs, simulating distant outcomes to avoid local optima while maintaining focus on actionable near-term steps. Crucially, these predictions incorporate uncertainty quantification \cite{haftner2020dream,assran2023self}, distinguishing predictable regularities (daily patterns) from stochastic events (sudden changes) to optimize risk-aware planning. The simulation prediction improves sample efficiency \cite{yang2023diffusion, wang2024worlddreamer,cho2024sora,li2025worldmodelbench} by replacing costly trial-and-error with mental rehearsal, particularly valuable in safety-critical domains like autonomous driving or robotic surgery. Furthermore, continuous prediction-error minimization drives iterative model refinement \cite{xiang2023language,mazzaglia2024genrl,gupta2024essential,jin2025embodied}, creating self-correcting systems that align their internal physics simulators with observed reality. Such anticipatory capabilities ultimately grant artificial agents human-like foresight, transforming reactive responses into purposeful, future-optimized behaviors.

\vspace{-0.5em}
\subsection{Classification of World Models for Embodied AI}
\label{CLASS_WMs}

Embodied AI with WMs can mainly be divided into three critical structures: the Recurrent State Space Model-based (RSSM-based) WMs for embodied AI, the Joint-Embedding Predictive Architecture-based (JEPA-based) WMs for embodied AI, and the Transformer-based WMs for embodied AI. Hierarchical-based WMs \cite{hansen2024hierarchical} and diffusion-based WMs \cite{alonso2024diffusion} are similar to other structures and are shown in Fig. \ref{fig:4_1}.

\subsubsection{RSSM-based WMs for Embodied AI}
RSSM constitutes the fundamental architecture underpinning the Dreamer algorithm family \cite{haftner2020dream, haftner2021mastering, okada2022dreamingv2, wu2023daydreamer}. This framework enhances predictive capabilities in latent representations by acquiring temporal environment dynamics through visual inputs, subsequently enabling action selection via latent trajectory optimization. Through orthogonal decomposition of hidden states into probabilistic and deterministic components, the architecture explicitly accounts for both systematic patterns and environmental uncertainties. Its demonstrated effectiveness in robotic motion control applications has inspired numerous derivative studies building upon its theoretical framework.

\subsubsection{JEPA-based WMs for Embodied AI}
JEPA \cite{lecun2022path} provides a structure for developing autonomous machine intelligence systems. This architecture establishes mapping relationships between input data and anticipated outcomes through representation learning. Diverging from conventional generative approaches, JEPA operates in abstract latent spaces rather than producing pixel-wise reconstructions, thereby prioritizing semantic feature extraction over low-level signal synthesis. A key methodological foundation of JEPA \cite{assran2023self} involves self-supervised training paradigms where neural networks learn to infer occluded or unobserved data segments. Such pre-training on extensive unlabeled datasets enables transfer learning across downstream applications, demonstrating enhanced generalization capabilities for both visual \cite{bardes2024revisiting, bardes2023mcjepa} and non-visual domains \cite{fei2024ajepa}.

\subsubsection{Transformer-based WMs for Embodied AI}
Originating in natural language processing research, the Transformer structure \cite{vaswani2017attention} fundamentally relies on attention mechanisms to process input sequences through parallelized context weighting. This design allows simultaneous computation of inter-element dependencies, overcoming the sequential processing constraints inherent in Recurrent Neural Networks (RNNs). Empirical evidence demonstrates superior performance in domains requiring persistent memory retention and explicit memory addressing for cognitive reasoning \cite{banino2020memo}, which has propelled its adoption in reinforcement learning research since 2020. Existing advancements have successfully implemented WMs using Transformer variants \cite{micheli2023transformers, robine2023transformer, wu2023paragraph}, outperforming RSSM architectures in memory-intensive interactive scenarios \cite{chen2022transdreamer}. Notably, Google's Genie framework \cite{bruce2024genie} employs the Spatial-Temporal Transformer (ST-Transformer) \cite{xu2021spatial} to create synthetic interactive environments through large-scale self-supervised video pretraining. This breakthrough establishes novel paradigms for actionable world modeling, revealing transformative potential for WMs development trajectories.

\section{Embodied AI with MLLMs and WMs}
\begin{figure*}[t]
\centering
\includegraphics[width=0.87\textwidth]{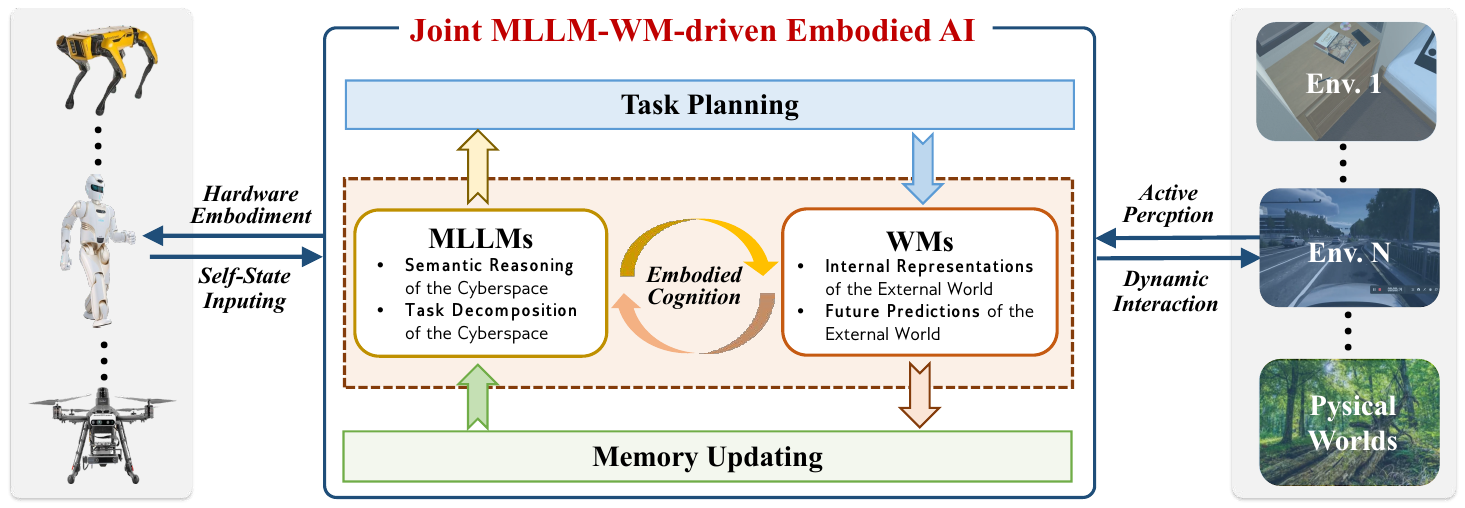}
\caption{Embodied AI with MLLMs and WMs. MLLMs can enhance WMs by injecting semantic knowledge for task decomposition and long-horizon reasoning, while WMs can assist MLLMs by building the physical world’s internal representations and future predictions, making joint MLLM-WM a promising architecture for embodied systems.}
\label{fig:5_1}
\vspace{-1em}
\end{figure*}

This section provides a comprehensive overview of embodied AI with MLLMs and WMs. We first elaborate in detail on the limitations of MLLMs and WMs for embodied AI and explain how MLLMs boost WMs reasoning, and how WMs boost MLLMs interaction in Subsection \ref{MLLM_WM}. Then we design a joint MLLM-WM-driven embodied AI architecture in Subsection \ref{MLLM_WM_A}. Finally, we discuss the advantages and challenges of new architecture in Subsection \ref{MLLM_WM_B}.

\vspace{-0.5em}
\subsection{MLLMs and WMs}
\label{MLLM_WM}
MLLMs enable contextual task reasoning but overlook physical constraints, while WMs excel at physics-aware simulation but lack high-level semantics. Their joint bridges semantic intelligence with grounded physical interaction.

\subsubsection{The Limitations of MLLMs for Embodied AI (without WMs)} MLLMs exhibit two critical limitations in embodied AI applications. First, they often fail to ground predictions \cite{liu2024physics} in physics-compliant dynamics, leading to impractical plans. For example, ignoring friction or material properties when manipulating objects may cause slippage or task failure. Second, their poor real-time adaptation to environmental feedback limits responsiveness \cite{driess2023vla}. While MLLMs excel at semantic task decomposition, they struggle to adaptively adjust actions when the environment changes dramatically. These limitations stem from their reliance on static, pre-trained knowledge rather than continuous physical interaction.

\subsubsection{The Limitations of WMs for Embodied AI (without LLMs/MLLMs)} WMs face limitations in abstract reasoning and generalization. They struggle with open-ended semantic tasks \cite{clark2013predictive} due to their focus on physical simulation rather than contextual understanding. Additionally, WMs lack generalizable task decomposition \cite{ha2018world} without explicit priors. For example, a WM model trained on rigid-object manipulation may fail to adapt to deformable materials without extensive retraining. Their predictive accuracy heavily depends on domain-specific interaction records, hindering scalability across diverse environments.

\subsubsection{MLLMs Boosting WMs Reasoning}
By leveraging cross-modal alignment and semantic grounding, MLLMs enable WMs to process complex environments dynamically, improving semantic reasoning, task decomposition, and human-robot interaction. 1) MLLMs can enrich WMs by fusing visual, auditory, and textual data into unified semantic representations. For instance, CLIP-based architectures \cite{radford2021learning} enable agents to align visual scenes with linguistic cues, reducing ambiguity in object recognition \cite{huang2023language}. 2) MLLMs can augment WM's task decomposition capacity by decomposing high-level goals into executable sub-tasks. Models like GPT-4V \cite{openai2023gpt4} generate step-by-step plans using environmental context stored in WM. For robotic manipulation, Code-as-Policies \cite{liang2023code} translates natural language instructions into code snippets, leveraging WM to track intermediate states. 3) MLLMs enable WMs to refine internal representations through human feedback. Techniques like Reinforcement Learning with Human Feedback (RLHF) \cite{ouyang2022training} allow agents to update WM priors based on corrective inputs \cite{shinn2023reflexion}. Those works in this Subsubsection are all possible ways for MLLMs to boost WMs reasoning, which is not achieved in existing works.

\subsubsection{WMs boosting MLLMs Interaction}
WMs can play a pivotal role in refining MLLMs by providing physical laws, spatio-temporal relationships, and closed-loop interaction experiences. WMs can mitigate MLLMs’ inherent limitations in temporal coherence and environmental grounding, enabling more robust decision-making in dynamic embodied tasks. 1) WMs can provide MLLMs with explicit representations of physical laws (e.g., gravity, friction) and commonsense rules to constrain action proposals. For example, Physion++ \cite{sun2023physics} integrating WM-stored biomechanical models can be used to filter MLLM-generated robotic motions violating torque limits; RoboGuide \cite{zhang2023slam} injects spatial occupancy maps into MLLM planners, preventing collisions during navigation. 2) WMs can stabilize MLLMs reasoning by maintaining spatio-temporal context during multimodal processing. For instance, MemPrompt \cite{xu2023contextual} can use WM buffers to align visual object trajectories with linguistic descriptions, resolving ambiguities in cluttered environments; RoboMem \cite{chen2023dynamic} can leverage WM-prioritized attention to filter irrelevant sensory noise, improving MLLM-based scene understanding. 3) WMs can enable iterative refinement of MLLM outputs through closed-loop interaction. Reflexion \cite{shinn2023reflexion} can store task-execution histories in WM, allowing MLLMs to correct kinematic errors using failure patterns \cite{liang2023code}. Those works in this Subsubsection are all possible ways for WMs to boost MLLMs' decisions, which has not been achieved in existing works.

\vspace{-0.5em}
\subsection{Joint MLLM-WM-driven Embodied AI Architecture}
\label{MLLM_WM_A}

We propose a joint MLLM-WM-driven embodied AI architecture (as shown in Fig. \ref{fig:5_1}), shedding light on their profound significance in enabling complex tasks within physical worlds. The specific workflow is as follows, with arrows highlighting the data exchange process.

\subsubsection{Robots → Self-State Inputing → MLLMs/WMs → Hardware Embodiment → Robots}
The process initiates with self-state inputting tracking proprioceptive metrics, such as degrees of freedom, number of sensors, etc. These metrics feed into both WMs and MLLMs: WMs use them to build internal representations of the agent’s physical state, while MLLMs contextualize these states for task alignment. Hardware embodiment is focused on implementing WMs and MLLMs into physical devices to solve sim-to-real problems. This bidirectional flow ensures actions respect both mechanical limits and high-level goals. 

\subsubsection{MLLMs → Task Planning → WMs → Memory Updating → MLLMs}  
MLLMs decompose abstract instructions into sub-tasks. A forward arrow delivers this plan to WMs, which predict outcomes based on existing environmental modeling. During execution, WMs log outcomes into memory. A vertical arrow transmits these logs to memory updating modules, which structure memory into experiences, represent the forgetting of past task memories, the renewal of current task memories, and the prediction of future task memories. These are then fed back to MLLMs via an arrow, enriching their knowledge base. This enables lifelong learning,  where past failures directly inform future planning.

\subsubsection{Environments → Active Perception → MLLMs/WMs → Dynamic Interaction → Environments}  
WMs first drive active perception by predicting key environmental changes. Multimodal inputs are then used to construct an internal representation of the external world through WMs and semantic reasoning through MLLMs. Then, the task decomposition of MLLMs and future prediction of WMs enable action selection and environmental interaction. Adaptive perception and interaction of dynamic environments are achieved through continuous iteration.

\definecolor{l}{RGB}{240,240,240}
\definecolor{h}{RGB}{255,255,200}
\definecolor{s}{RGB}{255,230,180}

\begin{table*}[ht]
\caption{Qualitative Comparison of MLLM-only, WM-only, and Joint MLLM-WM Architectures in Embodied AI. \colorbox{l}{Low}, \colorbox{h}{Medium}, \colorbox{s}{High}.}
\label{tab:my-table}
\renewcommand{\arraystretch}{1.5}
\resizebox{\textwidth}{!}{
\begin{tabular}{>{\centering}m{2cm} >{\centering\arraybackslash}m{3.7cm} >{\centering\arraybackslash}m{3.7cm} >{\centering\arraybackslash}m{4cm}}
\toprule
\textbf{Performance} & \textbf{LLM/MLLM-only} & \textbf{WM-only} & \textbf{Joint MLLM-WM} \\
\midrule
Semantic Understanding & \cellcolor{h} Advantages in contextual task reasoning and natural language understanding & \cellcolor{l} Limited in open-ended semantic understanding & \cellcolor{s} Combines high-level semantic abstraction with grounded contextual alignment \\
Task Decomposition & \cellcolor{h} Sequential logic enables sub-task planning via language prompts & \cellcolor{l} Lacks generalizable task decomposition mechanisms & \cellcolor{s} Semantic plans refined through physical feasibility via joint planning-execution loop \\
Physics Compliance & \cellcolor{l} Ignores physical constraints and dynamics in real-world interaction & \cellcolor{h} Physics-aware simulation with temporal consistency & \cellcolor{s} Enforces semantic-physical alignment for safe and executable plans \\
Future Prediction & \cellcolor{l} Lacks imagination-based reasoning & \cellcolor{h} Long-horizon multi-step prediction with uncertainty modeling &\cellcolor{s} Combines symbolic foresight and physically grounded imagination \\
Real-time Interaction & \cellcolor{l} Poor responsiveness to environmental feedback and significant reasoning latency & \cellcolor{h} Supports real-time predictive control via future state simulation & \cellcolor{s} Enables online adaptation through iterative plan refinement and memory updating \\
Memory Structure & \cellcolor{l} Sparse and unstructured memory & \cellcolor{h} Structured latent space encodes object dynamics and causal relationships & \cellcolor{s} Integrates semantic memory and world modeling for lifelong learning and reflection \\
Scalability & \cellcolor{l} Limited to pre-trained task space & \cellcolor{l} Poor transfer to unseen tasks without retraining & \cellcolor{h} Cross-task, cross-domain generalization through symbolic and sensorimotor synergy \\
\bottomrule
\end{tabular}
}
\end{table*}

\vspace{-0.5em}
\subsection{Discussions}
\label{MLLM_WM_B}
Joint MLLM-WM offer a promising architecture for embodied AI. As shown in TABLE \uppercase\expandafter{\romannumeral4}, MLLMs excel in semantic reasoning, enabling high-level task decomposition, contextual understanding, and adaptive planning by leveraging multimodal inputs. Meanwhile, WMs provide grounded, physics-based simulations of environments, ensuring actions align with real-world constraints. This synergy allows agents to balance abstract reasoning with real-time physical interactions, enhancing decision-making in dynamic settings. For instance, MLLMs can generate task plans while WMs validate feasibility, enabling iterative refinement. Additionally, joint architectures support cross-modal generalization, improving robustness in partially observable or novel scenarios by bridging symbolic knowledge and sensorimotor experiences.

The challenges of joint MLLM-WM-driven embodied AI architecture include 1) real-time synchronization between MLLMs’ high-latency semantic processing and WMs’ physics-based representation, often leading to delayed responses in dynamic environments; 2) semantic-physical misalignment, where MLLM-generated plans violate unmodeled physical constraints; and 3) scalable memory management, as continuous updates to WM’s internal states risk overwhelming MLLMs with irrelevant context. Additionally, training such systems requires vast multimodal datasets covering rare edge cases, while ensuring robustness against sensor noise and partial observability remains unsolved. These challenges need lightweight MLLMs inference, tighter feedback loops, and dynamic context-filtering mechanisms to minimize latency.
 
\section{Embodied AI Applications}
This section overviews the application of embodied AI in service robots, rescue robots, and other domains, highlighting trends in joint MLLMs and WMs to advance active perception, embodied cognition and dynamic interaction.

\vspace{-0.5em}
\subsection{Service Robotics}
Embodied AI is becoming an important technology in the service field. It helps service robots go beyond fixed rules and perform tasks in a flexible way using different types of information. Recent research \cite{li2023lot, shen2025detach} highlights its flexible applications across various fields. In domestic settings, systems such as RT-2~\cite{zitkovich2023rt} and SayCan~\cite{brohan2023can} combine language instructions with robot control, allowing robots to do tasks such as stacking dishes or cooking. Few-shot learning methods like AED~\cite{yeh2024aed} acquire new skills from limited demonstrations. In healthcare, robots with multiple types of input can help with reminders, rehabilitation, and companionship.~\cite{canal2021preferences, kachouie2014socially}. In public environments, platforms like Habitat~\cite{savva2019habitat} and RT-X~\cite{jang2022bc} support navigation and item delivery, even in changing environments, without needing special training for each task. This makes the system more general and useful in real life.
 
However, current approaches remain limited in handling long-horizon tasks. As illustrated in Fig.~\ref{fig:5_1}, the joint of WMs and MLLMs is emerging as a key strategy for enhancing the autonomy and long-term reasoning capabilities of service robots. The WM maintains an evolving environment model for planning and simulation, while the MLLM grounds commands like “clean up the living room” into adaptive subtasks. This collaboration supports flexible reasoning, goal adaptation, and robust real-world execution.

\vspace{-0.5em}
\subsection{Rescue UAVs}
Embodied AI technology technology is changing the way drones are used in disaster situations. Traditional drones are either manually controlled or rely on pre-built maps when in use, which leads to their inability to adapt to the environment independently. However, embodied drones \cite{feng2024multi, feng2025u2udata} can sense the environment in real time and respond to sudden changes. This ability makes them very useful in dangerous places like earthquake zones, forest fires, or floods. Recent studies show that embodied drones can perform many complex tasks. For instance, with the help of language models, they can understand and follow human voice instructions, helping drones quickly change their actions and enhancing their responses in emergency situations, such as “search near the collapsed bridge” \cite{zhang2024vision,an2024etpnav, flammini2022towards, chen2023human, lin2022event}. Secondly, some work use world models to simulate dangerous environments, which helps them avoid danger and plan a safer path \cite{pesonen2025detecting,javaid2024large, chen2024sdpl}. Other studies explore how multiple drones can work together to find survivors and map damaged areas \cite{zhang2024towards,zhou2022swarm, chen2024scale}.

However, despite these advancements, current approaches remain limited in handling long-horizon reasoning and autonomous decision-making under uncertainty. As illustrated in Fig.~\ref{fig:5_1}, jointing WMs and MLLMs has emerged as a key strategy for further enhancing UAV autonomy. The WM maintain a continuously evolving spatiotemporal representation of the environment, supporting planning and risk prediction even in GPS-denied conditions. The MLLM grounds commands into structured subtasks based on the UAV’s belief state. This coordination improves generalization, long-horizon reasoning, and high-level autonomy in mission-critical conditions.

\vspace{-0.5em}
\subsection{Industrial Robots}
Embodied AI is changing the way robots work in factories. With embodied AI, industrial robots \cite{zheng2025cross} can make smarter decisions based on their surroundings. Traditional industrial robots are usually fixed in one place. They use special sensors and tools and are required to complete tasks with very high accuracy. Because of this, they are better at doing jobs that need the same movements again and again.

However, with embodied AI, these robots can do more than repeat actions. By combining MLLMs and WMs, industrial robots can adjust how hard they hold fragile objects, or find a new path when they meet an obstacle. This has already been used in real life. For example, robots in Tesla’s factory can find and fix parts that are not lined up, without help from people.At JD, robots \cite{qin2022jd, bogue2024role} use different sensors to sort packages by size and address. In Tmall’s warehouse \cite{prawira2023robot}, robots use thermal cameras, LiDAR, and RGB sensors to check for problems in the inventory and send alerts when something is wrong. These examples show that embodied AI is helping robots become more flexible, reliable, and smart in factories.

\vspace{-0.5em}
\subsection{Other Applications}
In addition to its use in homes, healthcare, and rescue missions, embodied AI is also being applied in educational, virtual, and space environments \cite{liu2025tcdformer}. In smart manufacturing, it supports robots that can work together with humans, perform accurate assembly tasks, and adapt their actions based on changes in the workspace or human behavior \cite{feng2022timely, feng2019vabis, feng2025meta}. With the help of visual and touch feedback, these robots can safely handle fragile items~\cite{jin2024robotic,xue2025reactive}. In education, embodied AI is used in social robots that adjust their speech, gaze, and gestures according to the student’s focus and emotions \cite{YeWWXLS23, yeemotional, YeWWMH0S23}. This helps create a more personalized learning experience and builds long-term trust between students and robots~\cite{belpaeme2018social,lampropoulos2025social}. In virtual environments, embodied agents learn to move, interact with objects, and complete tasks that require several steps. They also develop memory over time to improve their performance~\cite{wang2023voyager}. In space exploration, where conditions are unknown and communication with Earth is delayed, embodied AI allows robots to make decisions on their own and adapt to new surroundings~\cite{qunzhi2024research}. These examples show that embodied AI is becoming more flexible and useful across many fields, helping machines see, act, and learn in both real and virtual worlds.

\section{Future Directions}
As embodied AI moves from simulation to real-world deployment, future research must prioritize the development of unified and reliable systems across several core domains. Key directions include autonomous embodied AI, embodied AI hardware, swarm embodied AI, and evaluation benchmark. 

\subsection{Autonomous Embodied AI}

The purpose of autonomous embodied AI is to enable agents to operate independently for a long time in a dynamic and open environment. Future research is expected to develop along several key directions. First, adaptive perception can give the system the ability to autonomously select input data, which can be achieved by dynamically choosing and integrating information from different sensory modalities. Second, Building on this foundation, building environmental awareness is essential. Environmental awareness helps agents quickly adapt to changes, predict the consequences of their actions, and transfer their behavior to new environments. It requires memory architectures that can capture spatiotemporal patterns and model causal relationships. Third, future systems should combine MLLMs with real-time physical interaction, which allows agents to bridge high-level language instructions with low-level control, and accurately model the real physical world.

\vspace{-0.5em}
\subsection{Embodied AI Hardware}
Future research in embodied AI hardware is expected to advance in the following four directions.  First, hardware-aware model compression will continue to integrate techniques such as quantization and pruning with hardware performance metrics, enabling precise control over the trade-off between model accuracy and deployment efficiency.  Second, graph-level compilation optimization will play a key role in bridging the gap between high-level embodied models and low-level hardware execution, which will focus on more effective operator fusion, scheduling strategies, and memory access efficiency to reduce execution overhead.  Third, domain-specific accelerators will be increasingly tailored to the computational characteristics of embodied tasks. Reconfigurable architectures such as FPGA and CGRA offer flexibility and adaptability, while ASIC-based designs provide high efficiency and performance.  Fourth, hardware-software co-design will become essential for eliminating mismatches between algorithm behavior and hardware architecture.  Joint optimization of model structures and hardware architecture will be critical to achieving real-time, energy-efficient execution in embodied systems.

\vspace{-0.5em}
\subsection{Swarm Embodied AI}
Swarm embodied AI refers to the collaborative perception and decision-making of multiple agents. refers to the collaborative perception and decision-making of multiple agents. Because multiple agents can exhibit stronger capabilities when cooperating than a single agent, this kind of "collective intelligence" has aroused the interest of many researchers and is also regarded as an important step for agents to approach humans. First of all, to enable multiple agents to cooperate smoothly, it is necessary to develop collaborative WMs. This model can establish a shared and dynamic environmental representation based on the observations of each agent, forming the basis of collective understanding. Secondly, multi-agent representation learning is very important. It can help the agent understand its own state and also comprehend the situations of other agents. This is the basis for communication and cooperation among agents. In addition, modeling social behavior among agents is also crucial. Role allocation and group decision-making can be better achieved through behavioral modeling. Finally, to seamlessly integrate into real-world applications, it is also important to design natural human-swarm interaction interfaces. It may include multimodal language foundations and get-based control methods, making it easier for humans to direct and guide the entire agent group.

\subsection{Explainability and Trustworthiness Embodied AI}
Explainability and trustworthiness represent a critical frontier for Embodied AI, essential for its safe, ethical, and widespread real-world deployment as agents increasingly interact physically with humans and dynamic environments. Future research must address several key challenges: Firstly, designing benchmarks that provide real-time, human-understandable justifications for agent actions, particularly during unexpected situations or failures, is crucial for user trust and debugging. Secondly, establishing robust mechanisms to ensure agents adhere to ethical principles and human values during autonomous decision-making, especially in morally ambiguous scenarios common in rescue or healthcare applications, requires significant advancement. Thirdly, creating verifiable safety guarantees and certification standards for agents operating in unstructured physical settings, mitigating risks associated with unpredictable interactions, remains an open problem. Finally, enhancing robustness against adversarial attacks, sensor noise, and distribution shifts, ensuring reliable performance despite uncertainties inherent in the real world, is fundamental for trustworthy operation. Addressing these multifaceted research problems in explainability and trustworthiness is paramount, as progress in this direction will unlock the full potential of Embodied AI by fostering user confidence, enabling responsible innovation, and facilitating regulatory acceptance.

\vspace{-0.5em}
\subsection{Other Directions}
Several new directions may influence the future development of embodied AI. One important direction is lifelong learning. Agents need to continuously learn new skills without forgetting what they have already learned. Only in this way can they adapt to the dynamic environment and maintain the accuracy of the previously completed tasks. Another key direction is human-in-the-loop learning. Human feedback is very important supervisory information. A small amount of feedback can significantly improve the performance of an agent and make it more human-like. To achieve this goal, we need better methods to enable agents to understand human goals and preferences. Finally, as agents become more autonomous, moral decision-making becomes increasingly important. Future systems should learn to carefully identify moral hazard and follow human values. This will help ensure that the embedded artificial intelligence is both safe and reliable.

\section*{Acknowledgement}
This work is supported by National Natural Science Foundation of China No.62222209, Beijing National Research Center for Information Science and Technology under Grant No.BNR2023TD03006, China Postdoctoral Science Foundation under Grant No.2024M751688, Postdoctoral Fellowship Program of CPSF under Grant No.GZC20240827.

We thank Dr. Ren Wang and Dr. Mingzi Wang for their contribution to Section \uppercase\expandafter{\romannumeral2}. We thank Dr. Yu-Wei Zhan for her contributions to Section \uppercase\expandafter{\romannumeral6} and Section \uppercase\expandafter{\romannumeral7}.

\bibliographystyle{./IEEEtran}
\bibliography{./IEEEabrv, ./EAI} 

\vspace{-1em}
\begin{IEEEbiography}[{\includegraphics[width=1in,height=1.25in,clip,keepaspectratio]{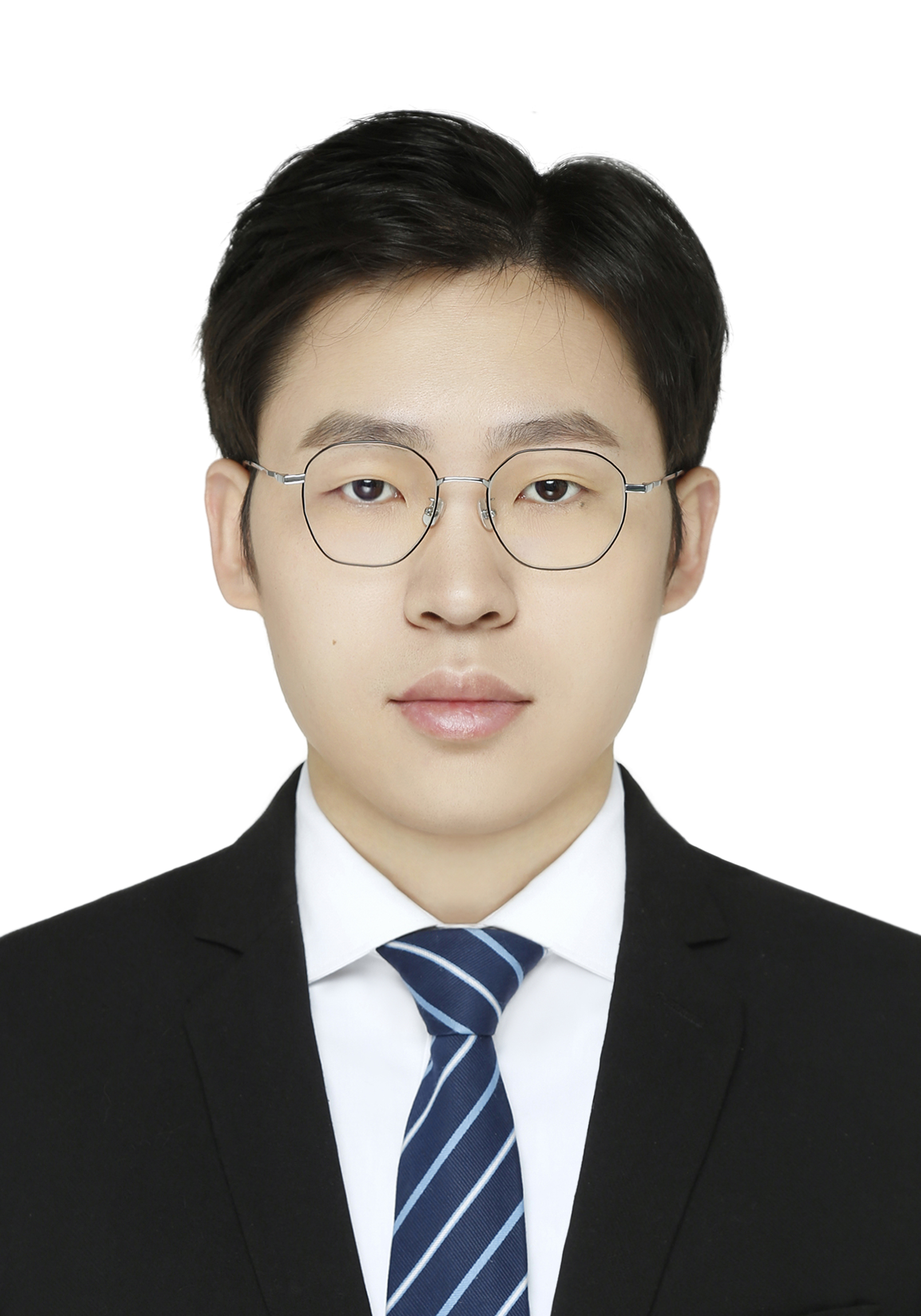}}]{Tongtong Feng} is currently a postdoctoral researcher at the Department of Computer Science and Technology, Tsinghua University. He got his Ph.D. degree in Computer Science and Technology from Beijing University of Posts and Telecommunications. His research interests include Embodied AI, World Model, and Multimedia Intelligence. He has published over 15 high-quality research papers in top journals and conferences, including IEEE TMM, ESWA, ACM Multimedia and AAAI, etc. He got the Best Paper Nomination of ACM Multimedia 2024.
\end{IEEEbiography}

\vspace{-1em}

\begin{IEEEbiography}[{\includegraphics[width=1in,height=1.25in,clip,keepaspectratio]{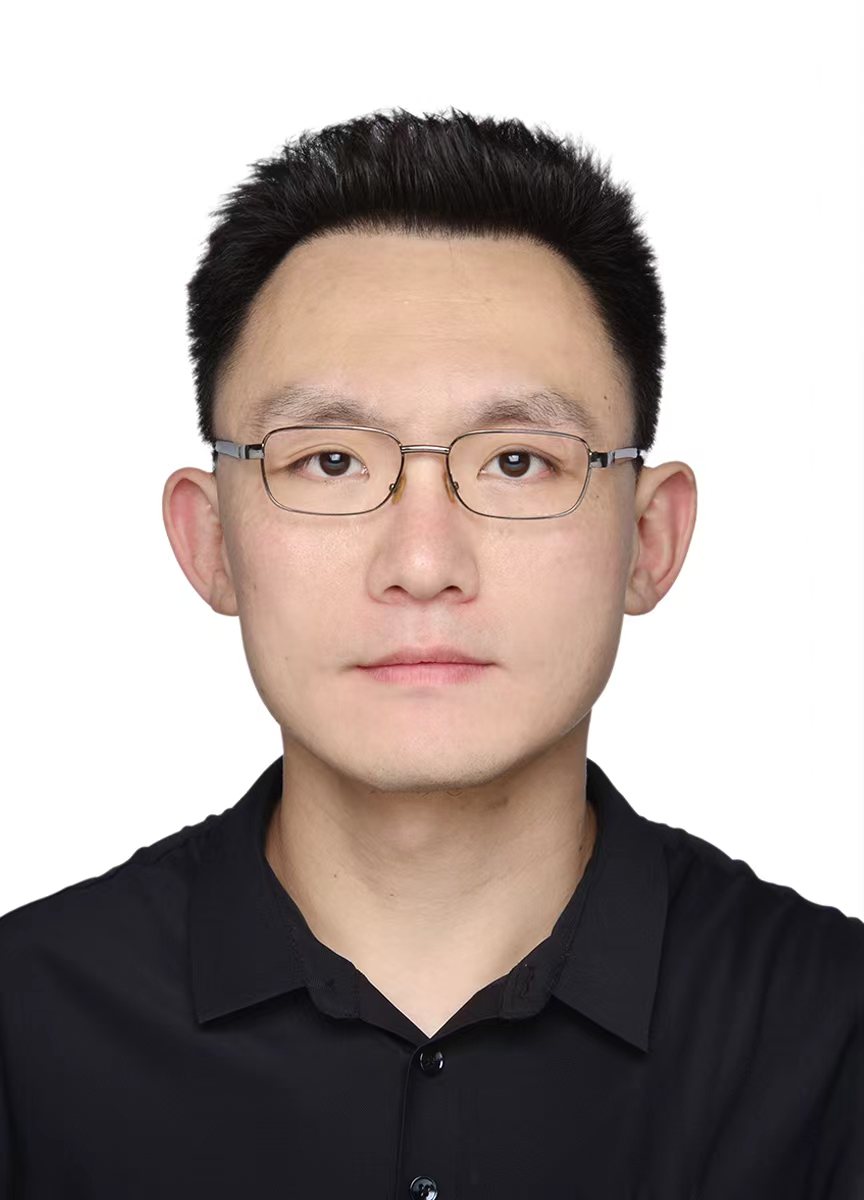}}]{Xin Wang} is currently an Associate Professor at the Department of Computer Science and Technology, Tsinghua University. He got both his Ph.D. and B.E degrees in Computer Science and Technology from Zhejiang University, China. He also holds a Ph.D. degree in Computing Science from Simon Fraser University, Canada. His research interests include multimedia intelligence, machine learning and its applications in multimedia big data analysis. He has published over 150 high-quality research papers in top journals and conferences including IEEE TPAMI, IEEE TKDE, ACM TOIS, ICML, NeurIPS, ACM KDD, ACM Web Conference, ACM SIGIR and ACM Multimedia etc., winning three best paper awards. He is the recipient of 2020 ACM China Rising Star Award, 2022 IEEE TCMC Rising Star Award and 2023 DAMO Academy Young Fellow.
\end{IEEEbiography}

\vspace{-1em}

\begin{IEEEbiography}[{\includegraphics[width=1in,height=1.25in,clip,keepaspectratio]{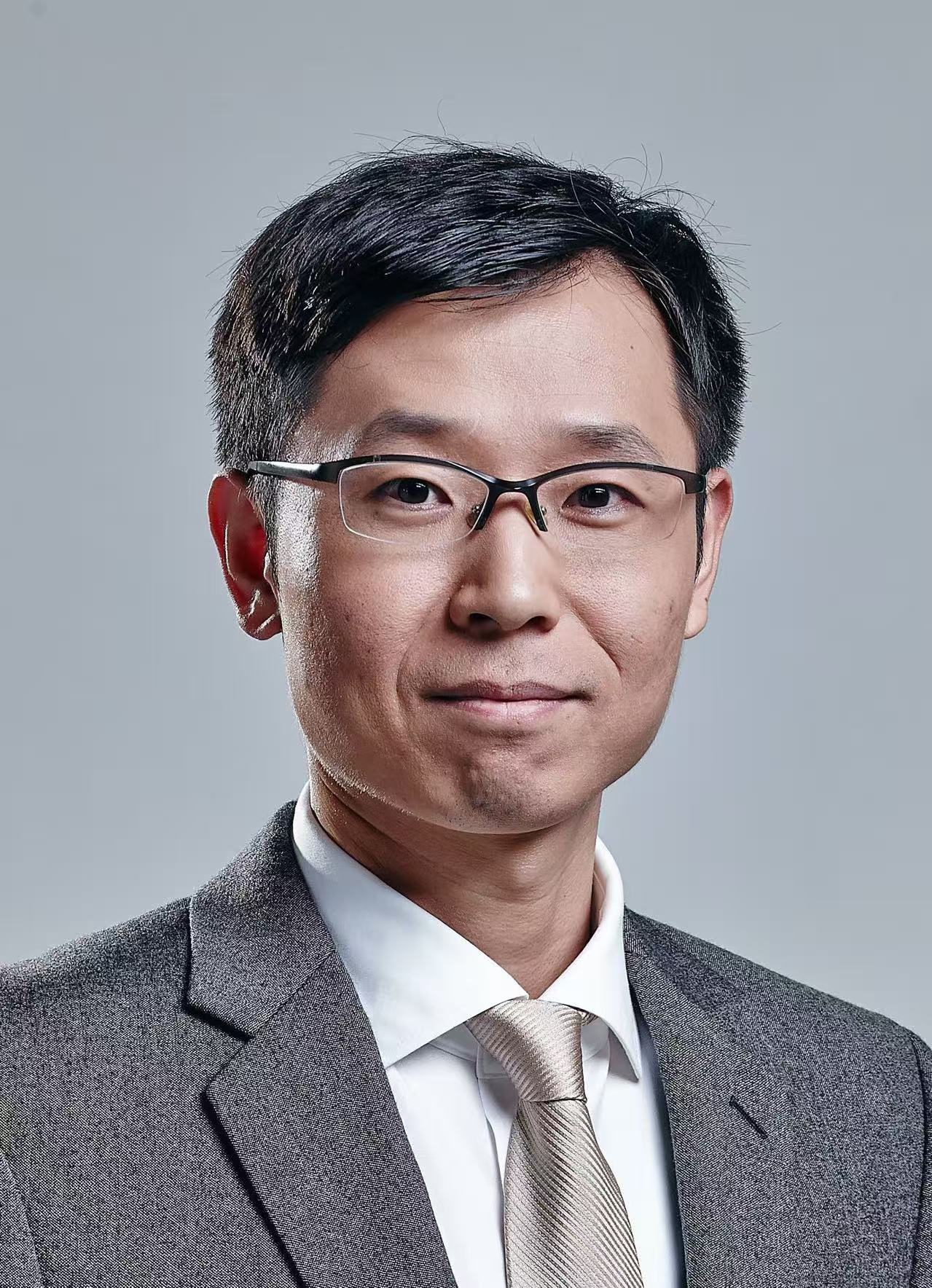}}]{Yu-Gang Jiang} (Fellow, IEEE) received the PhD degree in Computer Science from City University of Hong Kong in 2009 and worked as a Postdoctoral Research Scientist at Columbia University, New York, during 2009-2011. He is currently a Distinguished Professor of Computer Science at Fudan University, Shanghai, China. His research lies in the areas of multimedia, computer vision, embodied AI and trustworthy AI. His research has led to the development of innovative AI tools that have been used in many practical applications like defect detection for high-speed railway infrastructures. His open-source video analysis toolkits and datasets such as CU-VIREO374, CCV, THUMOS, FCVID and WildDeepfake have been widely used in both academia and industry. He currently serves as Chair of ACM Shanghai Chapter and Associate Editor of several international journals. For contributions to large-scale and trustworthy video analysis, he was elected to Fellow of IEEE, IAPR and CCF.
\end{IEEEbiography}

\vspace{-1em}

\begin{IEEEbiography}[{\includegraphics[width=1in,height=1.25in,clip,keepaspectratio]{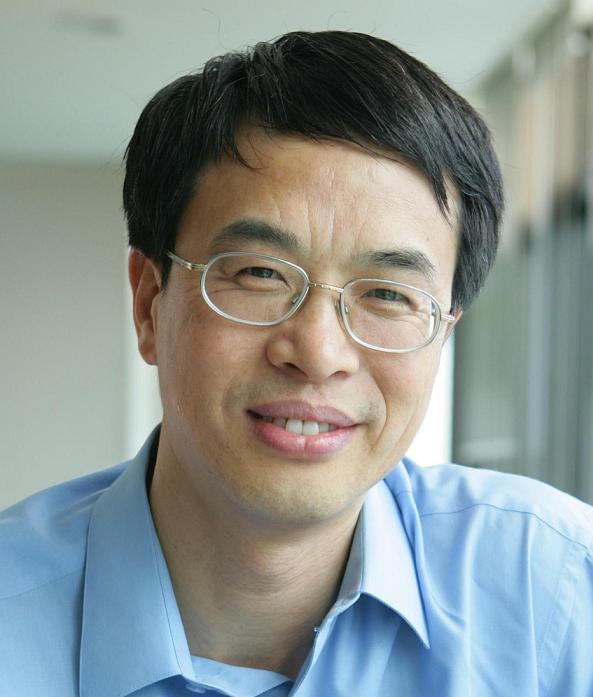}}]{Wenwu Zhu} is currently a Professor in the Department of Computer Science and Technology at Tsinghua University, the Vice Dean of Beijing National Research Center for Information Science and Technology. Prior to his current post, he was a Senior Researcher and Research Manager at Microsoft Research Asia. He was the Chief Scientist and Director at Intel Research China from 2004 to 2008. He worked at Bell Labs New Jersey as Member of Technical Staff during 1996-1999. He received his Ph.D. degree from New York University in 1996.

His research interests include graph machine learning, curriculum learning, data-driven multimedia, big data. He has published over 400 referred papers, and is inventor of over 100 patents. He received ten Best Paper Awards, including ACM Multimedia 2012 and IEEE Transactions on Circuits and Systems for Video Technology in 2001 and 2019.  

He serves as the EiC for IEEE Transactions on Circuits and Systems for Video Technology (2024-2025), the EiC for IEEE Transactions on Multimedia (2017-2019) and the Chair of the steering committee for IEEE Transactions on Multimedia (2020-2022). He serves as General Co-Chair for ACM Multimedia 2018 and ACM CIKM 2019. He is an AAAS Fellow, IEEE Fellow, ACM Fellow, SPIE Fellow, and a member of Academia Europaea.
\end{IEEEbiography}

\end{document}